\documentclass[10pt,journal,final,twocolumn]{IEEEtran}
\ifCLASSOPTIONcompsoc
\else
\fi

\ifCLASSINFOpdf
\else
\fi

\ifCLASSOPTIONcompsoc
  \usepackage[nocompress]{cite}
\else
  \usepackage{cite}
\fi

\usepackage[cmex10]{amsmath}
\usepackage{mathtools}
\usepackage{algorithmic}
\usepackage{epsfig}
\usepackage{graphicx}
\usepackage{subfigure}
\usepackage{amssymb}
\usepackage{afterpage}
\usepackage{url}

\begin{document}
\title{Linear Support Tensor Machine: \\Pedestrian Detection in Thermal Infrared Images}
\author{Sujoy~Kumar~Biswas,~\IEEEmembership{Student~Member,~IEEE,}
        Peyman~Milanfar,~\IEEEmembership{Fellow,~IEEE}}

\markboth{IEEE Transactions on Image Processing (Submitted)}%
{Biswas \MakeLowercase{\textit{et al.}}}
\maketitle
\IEEEdisplaynotcompsoctitleabstractindextext
\IEEEpeerreviewmaketitle

\begin{abstract}
Pedestrian detection in thermal infrared images poses unique challenges because of the low resolution and noisy nature of the image. Here we propose a mid-level attribute in the form of multidimensional template, or tensor, using Local Steering Kernel (LSK) as low-level descriptors for detecting pedestrians in far infrared images. LSK is specifically designed to deal with intrinsic image noise and pixel level uncertainty by capturing local image \emph{geometry} succinctly instead of collecting local orientation \emph{statistics} (e.g., histograms in HOG). Our second contribution is the introduction of a new image similarity kernel in the popular maximum margin framework of support vector machines that results in a relatively short and simple training phase for building a rigid pedestrian detector. Our third contribution is to replace the sluggish but \emph{de facto} sliding window based detection methodology with multichannel discrete Fourier transform, facilitating very fast and efficient pedestrian localization. The experimental studies on publicly available thermal infrared images justify our proposals and model assumptions. In addition, the proposed work also involves the release of our in-house annotations of pedestrians in more than 17000 frames of OSU Color Thermal database for the purpose of sharing with the research community.
\end{abstract}

\section{Introduction}
The computer vision community has made good progress in people and pedestrian detection in natural images and videos in the last ten years \cite{dalal2005histograms, felzenszwalb2010object, dollar2012pedestrian, ouyang2013joint, benenson2014ten}. However, such endeavors in locating pedestrians have mostly been restricted to photographs captured with visible range sensors. Infrared and thermal imaging sensors, which provide excellent visible cues in unconventional settings (e.g., night time visibility), have historically found their use limited to military, security and medical applications. However, with increasing image quality and decreasing price and size, some of the thermal sensing devices are finding commercial deployment for home and office monitoring as well as automotive applications \cite{geronimo2010survey, hotspot14}. Research effort has so far been limited in this domain for building reliable and efficient computer vision systems for infrared thermal image sensors. The objective of this paper is to address this concern.

\begin{figure*}[!t]
\begin{center}
\subfigure[]{{}\includegraphics[width=0.24\linewidth]{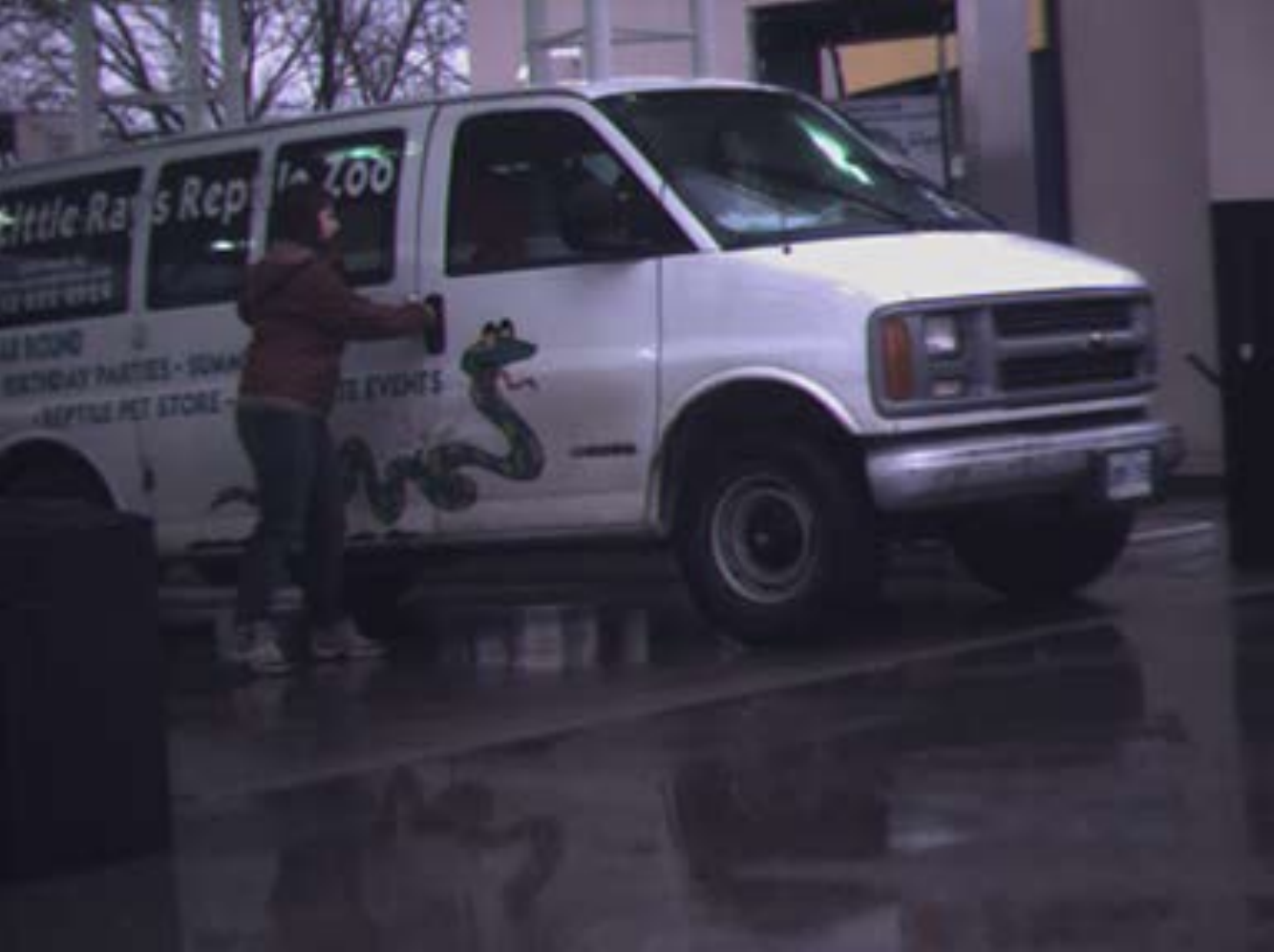}}
\subfigure[]{{}\includegraphics[width=0.24\linewidth]{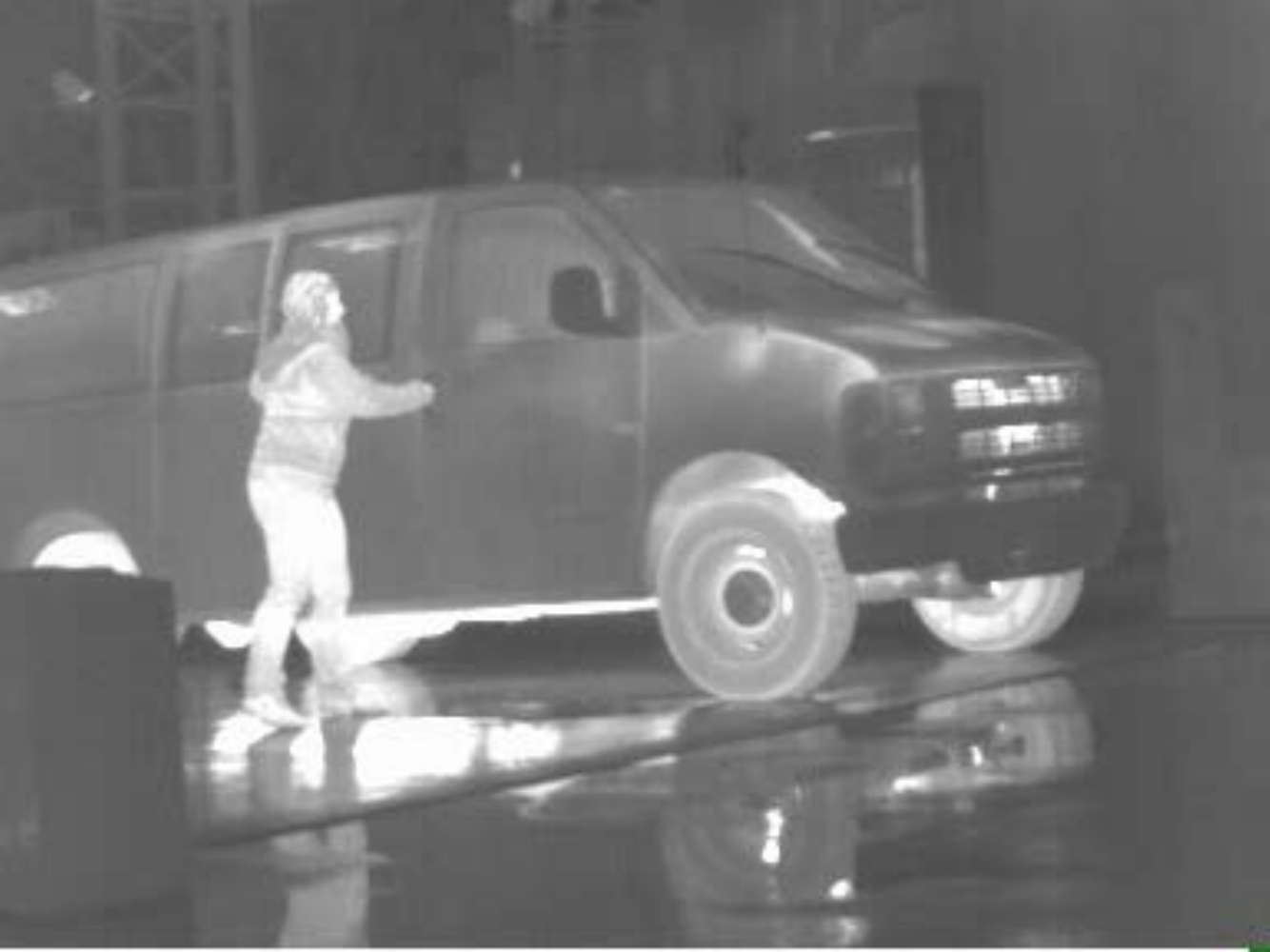}}
\subfigure[]{{}\includegraphics[width=0.24\linewidth]{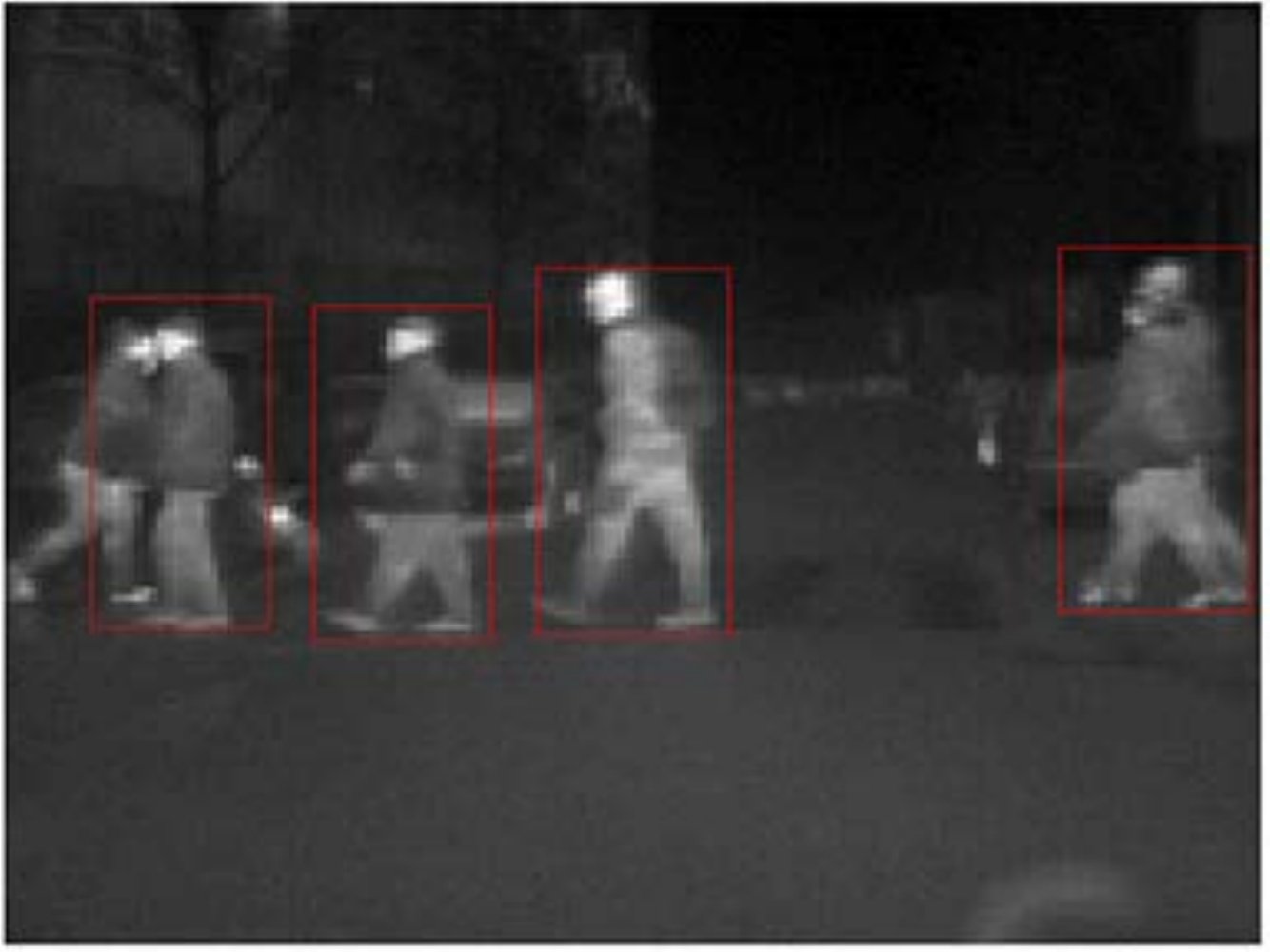}}
\subfigure[]{{}\includegraphics[width=0.24\linewidth]{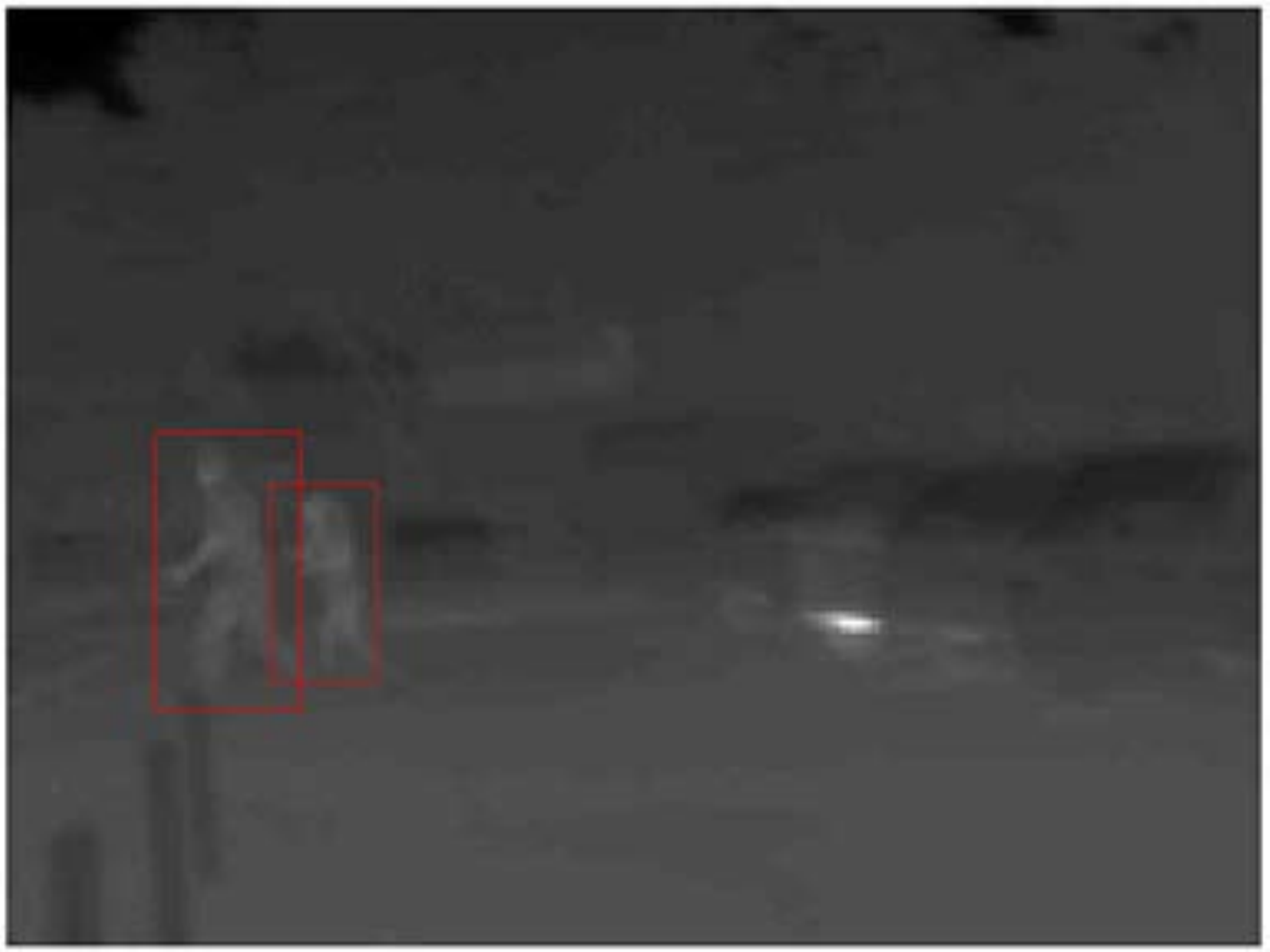}}
\end{center}
   \caption[]{Infrared images are different: natural color images exhibit textures which are suppressed in infrared images (left pair images \cite{morris2007statistics} \footnotemark[1]). As a consequence, many background texture features like trees and buildings may remain relatively nondescriptive (third from left) which complicates the separation of the background in feature space during the learning process. In addition, the high noise adds to the complexity of detecting foreground objects (far right).}
\label{F0}
\end{figure*}

Thermal image sensors typically have a spectral sensitivity ranging from 7 micron to 14 micron band of wavelength. The capacity of these imaging devices to appropriately capture images of objects depends on their emissivity and reflectivity in a nontrivial fashion. The material and surface properties of the objects control emissivity whereas amount of background radiation reflected by the objects influence their reflectivity. The involvement of multiple factors in the image formation process often leads to various distortions in thermal images, notably, \emph{halo effect, hotspot areas, radiometric distortions} to name a few \cite{goodall2016tasking}. Fig. \ref{F0} illustrates halo effect in a natural-thermal image pair. Also noticeable is the fact that textures visible on objects often get suppressed in thermal images. This fact has important bearing as far as visual recognition is concerned because the negative examples corresponding to the background tend to be far less descriptive. Fig. \ref{F0} also illustrates the challenge involved in detecting foreground objects because of inherently noisy nature of the  infrared images. From the representative images it is fair to conclude that a successful visual recognition system must include a strong measure of visual similarity that can overcome the effect of weak and ambiguous image signal as well as a robust noise handling component in the feature computation process.

In this paper we focus our attention on detecting pedestrians, particularly walking at a distance from the camera. Since our primary motivation for studying infrared images comes from automotive applications, we focus our attention and study on building pedestrian detectors without considering tracking information and background model. Of course, the proposed methodology is general enough to include such information toward building more sophisticated models. Ensemble based techniques like boosting and random forest \cite{dollar2014}, convolutional neural network (CNN) and deformable part model \cite{felzenszwalb2010object} are three widely used approaches toward building an effective object detector. Though extremely fast in runtime, training with boosting and CNN often takes too long to converge, sometimes spanning days. The ready availability of a substantially large clean annotation set is often recommended for feature learning particularly with deep architectures. In this paper, we shall not delve into feature learning (as done in CNN \cite{girshick2014rcnn}) but focus on a fast and efficient mid-level attribute that can allow clean, simple training phase with reasonably good detection performance. The widely successful pedestrian, and in general, pose detector deformable part model is built upon the fundamental notion of representing templates with the Histogram of Oriented Gradients (HOG) as mid-level representation. However, the runtime costs associated with the overhead of the \emph{part} complexities is a major issue in the part based detectors. Also, the small size of the pedestrians when they appear far from from the camera does not leave room for the explicit modeling of the parts/limbs. Interestingly, a recent study \cite{benenson2013seeking} has shown that with careful design a seemingly naive rigid detector can perform exceptionally well in comparison to its advanced counterparts with higher complexities. Our work draws inspiration from their study, and in this paper we revisit the simple but very effective detector of Dalal and Triggs \cite{dalal2005histograms}, with the following contributions.

{\bf Rigid template with LSK tensors}: dealing with heavy noise and artifacts in image signal while performing visual recognition has garnered relatively low attention from the community. This is particularly relevant in infrared domain where sensor noise is high, and feature variability is much less compared to natural photographs. Our objective is to capture local image structure in a stable and reliable fashion. For accomplishing that purpose we advocate the use of Local Steering Kernel (LSK)  \cite{Takeda2007Kernel, Seo2010training, Seo2011face} as low level image region descriptor. LSK had its genesis primarily in the image denoising and filtering tasks \cite{Takeda2007Kernel}, and is also known as Locally Adaptive Regression Kernel, or LARK, following the fact that LARK filter coefficients are computed adaptively following a local regression on neighboring pixel intensities. Using LSK \footnotemark[2] as low level descriptor we propose a tensor representation of our mid-level attribute --- the detector template. \footnotetext[1]{The left pair of images \cite{morris2007statistics} (available online, 5th August, 2016) is downloaded from: \url{http://www.dgp.toronto.edu/~nmorris/IR/} for academic use} \footnotetext[2]{We shall follow the name LSK since it intuitively indicates characteristics of the features} However in doing so, the geometric invariance (as in HOG \cite{dalal2005histograms, felzenszwalb2010object}) and scale invariance (as in SIFT \cite{lowe2004distinctive}) are relaxed at the expense of robustness of the descriptor. HOG and LSK both capture local orientation information. However, HOG computes orientation \emph{statistics} over a set of angular directions in a small spatial neighborhood, whereas LSK captures dominant local orientation and is thus more stable in dealing with image noise.

{\bf Maximum margin matrix cosine similarity}: our second contribution is a maximum margin learning methodology that respects and leverages the tensor form of our mid-level representation. Past work has highlighted the effective use of Matrix Cosine Similarity (MCS) as a robust measure for computing image similarity in a training-free, one shot detection scenarios. \cite{Seo2010training, biswas2016one}. Motivated by such findigs we have extended MCS to introduce a maximum margin training formulation for learning a decision boundary that can separate pedestrian from the background.

{\bf Beyond sliding window search}: the standard technique for object search, i.e., sliding window based object detection, incurs prohibitive computational cost. To resolve this issues Lampert \emph{et al.} have proposed a branch and bound technique \cite{lampertcvpr2008, lamperttpami2009}. In our work, we propose a relatively simple but efficient technique to improve the detection time, performing multiscale pedestrian detection in less than a second. The search for pedestrian in a test image proceeds in the frequency domain (using Fourier transform) with integral image based normalization, yielding an elegant framework for extremely efficient and fast pedestrian detection.

{\bf Analysis and Annotations}: we have demonstrated the efficacy of the proposed methodology on three standard benchmark datasets \cite{davis2005two, davis2007background, LSI}. In particular, we have annotated OSU Color-Thermal pedestrian dataset as we could not find a good annotation in the public domain suitable for evaluating various object detection algorithms. To help push the state of the art in this area our work also includes the ground truth annotations of pedestrians in 17088 frames in this dataset\footnotemark[3].\footnotetext[3]{Available for download from the first author's website}

The related work till date explored LSK in various detection scenarios on natural images but stayed limited on two counts. All such past studies i) did not explore or leverage inherent tensor connection of LSK, and ii) did not extend MCS toward a more general, learning based scheme. For example, Seo et al. \cite{Seo2010training, seo2011action} and Biswas et al. \cite{biswas2016one} restricted LSK and MCS to the study of training-free, generic one shot object detection. Subsequent investigations by Zoidi et al. \cite{zoidi2013visual} reported performance of LSK to detect humans in videos. In a further extension, You et al. \cite{you2014local} used LSK features for learning local metric for ensemble based object detection. Though laudable, in all of such research endeavors, the generalization principle of MCS had been missing. We believe it promising to show that MCS could be effectively and efficiently integrated with LSK for building large scale learning systems. The whole premise behind this work is that tensor representations when combined with MCS kernel would invariably lead to a reasonably simple and fast training scheme, with rapid and precise localization result.

We proceed with the system overview in the next section.

\section{System Overview}
Unlike color images that contain multiple color channels infrared images typically have a single channel that we denote by $M\times N$ image matrix ${\bf I}$, defined on a rectangular grid $\Omega \subset \mathbb{R}^2$. We densely compute low-level, $l$-dimensional descriptors ${\bf h}_i \in \mathbb{R}^l$, at each pixel location ${\bf x}_i \in \Omega$. Aggregating all `${\bf h}_i$'s together, we form our third-order descriptor tensor ${\bf H}\in \mathbb{R}^{M\times N \times l}$ corresponding to image ${\bf I}$. The order, or dimension, three in ${\bf H}$ alludes to the $l$ channels of the computed descriptor.

Dense computation makes the descriptor highly descriptive no doubt, but at the same time it invites the undue effect of redundancy. To distill the redundancy we reduce the number of channels in ${\bf H}$ from $l$ to $d$ by employing principal component analysis. The result is the decorrelated feature tensor ${\bf F}\in \mathbb{R}^{M\times N \times d}$, where $d << l$.

From $M\times N \times d$ tensor ${\bf F}$ we crop a smaller third order tensor window $m\times n \times d$ corresponding to the ground truth annotation that represents a `pedestrian'. We follow the similar process for collecting the negative examples that correspond to the `background'. Specifically, the $i$-th training example ${\bf F}_i \in \mathbb{R}^{m\times n \times d}$ associated with the class label $y_i$, is first normalized and then used as an input to a maximum margin classification using our proposed kernel function. The objective is to learn a decision boundary to separate pedestrians from background in the tensor feature space. We describe our full methodology starting with the feature computation in the next section.
\begin{figure}
\begin{center}
\includegraphics[width=0.5\textwidth]{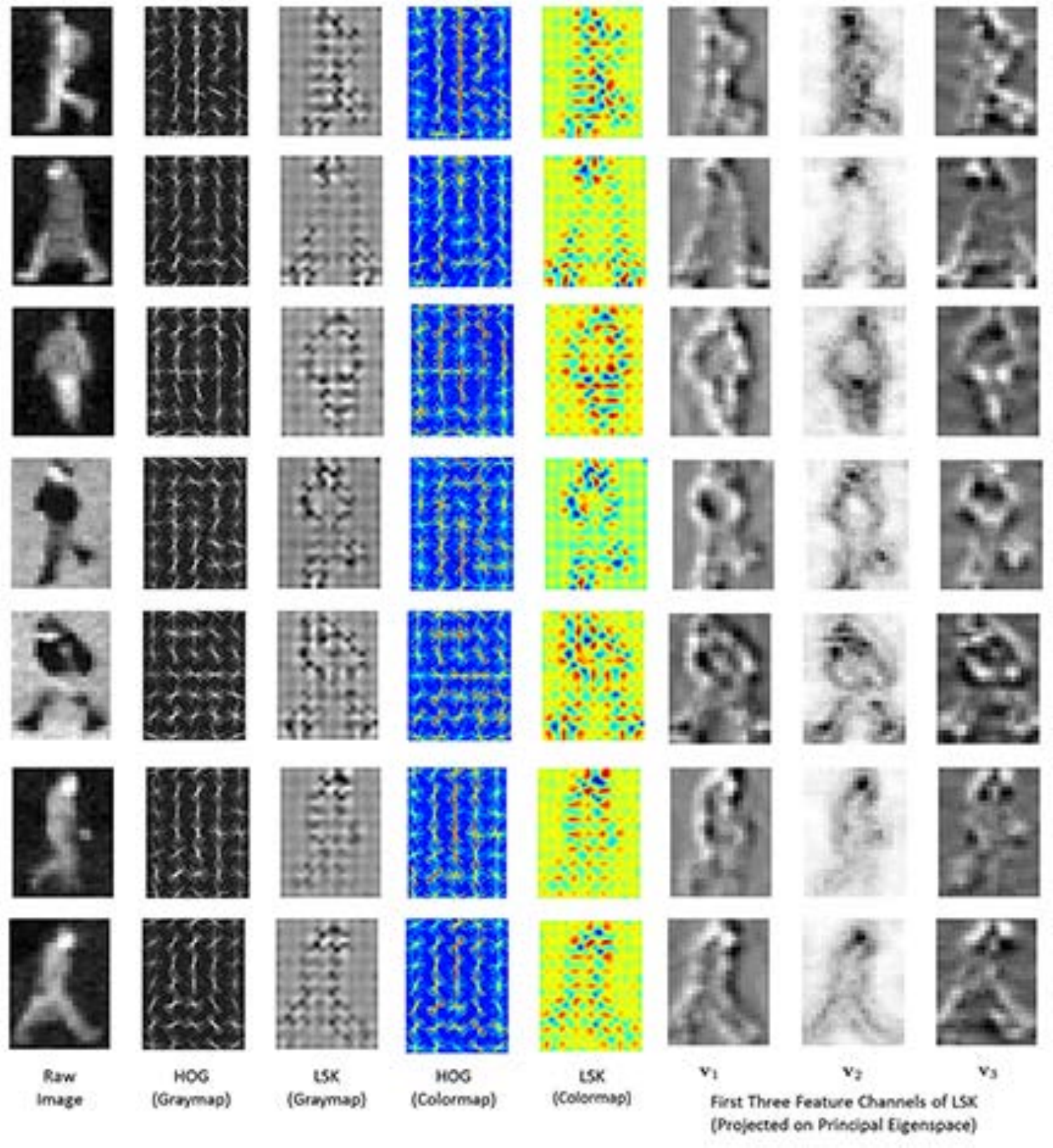}
\end{center}
\caption{{\bf LSK Visualization} First column displays raw infrared images of pedestrians in different poses. HOG and LSK features are displayed in grayscale (second and third column respectively) as well as in in colormap (fourth and fifth column respectively). LSK is displayed thus after computing them in non overlapping blocks. Columns sixth, seventh and eighth show LSK features after projecting LSK descriptors on three leading principal components.}
\label{F1}
\end{figure}
\section{Local Structure Estimation with Steering Kernel}
We represent each pixel by a two dimensional coordinate vector ${\bf x}_i = \left[ x_{i1}~ x_{i2} \right]' \in \Omega$. We define the image ${\bf I}$ as a function such that ${\bf I}: \Omega \rightarrow \mathbb{R}$. The value of image ${\bf I}$ at a particular pixel location ${\bf x}_i \in \Omega$ is given by the pixel intensity ${\bf I}({\bf x}_i)$.

LSK derives its name as well as much of its descriptive power from a steering matrix \cite{Takeda2007Kernel} (also known as gradient covariance matrix or structure tensor) that lies at its heart, and defined at the pixel ${\bf x}_i$ as follows:
\begin{equation}
\label{Cmatrix}
{\bf C}_{\Omega_i} = \sum_{{\bf x}_i \in \Omega_{i}} \begin{bmatrix} {\frac{\partial {\bf I}({\bf x}_i)}{\partial x_{i1}}}^2 & \frac{\partial {\bf I}({\bf x}_i)}{\partial x_{i1}} \cdot \frac{\partial {\bf I}({\bf x}_i)}{\partial x_{i2}} \\
\frac{\partial {\bf I}({\bf x}_i)}{\partial x_{i1}} \cdot \frac{\partial {\bf I}({\bf x}_i)}{\partial x_{i2}} & {\frac{\partial {\bf I}({\bf x}_i)}{\partial x_{i2}}}^2 \end{bmatrix},
\end{equation}
where $\Omega_i$ is the rectangular window centered at ${\bf x}_i$. In theory, the steering matrix is based on gradients $\frac{\partial {\bf I}({\bf x}_i)}{\partial {\bf x}}$ in a single pixel ${\bf x}_i$ \cite{Seo2011face}. However, a single pixel estimate makes the steering matrix unstable and prone to noisy perturbation of the data. Therefore, ${\bf C}_{\Omega_i}$ is the regularized estimate of the steering matrix which is \emph{averaged} (note the summation in Eq. (\ref{Cmatrix}) from a local aggregation of gradients over a rectangular window $\Omega_i$.

As the name suggests, the steering matrix captures the principal directions of local texture from the gradient distribution in the the small neighborhood $\Omega$ (mostly 5$\times$5). This idea becomes easy to follow if the spectral decomposition of the steering matrix is brought into picture:
\begin{equation}
\label{Cspectral}
{\bf C}_{\Omega_i} = \lambda_1 {\bf u}_1{\bf u}_1' + \lambda_2 {\bf u}_2{\bf u}_2',
\end{equation}
where, $\lambda_1, \lambda_2$ are the eigenvalues, and ${\bf u}_1, {\bf u}_2$ are the eignevectors representing principal directions. Denoting singular values as $s_1 = \sqrt{\lambda_1}$ and $s_2 = \sqrt{\lambda_2}$, they are turned into a Riemannian metric by the following regularization (to avoid numerical instabilities) while keeping the eigenvectors unaltered:
\begin{equation}
\label{Creg}
{\bf C}_{\Omega_i} = (s_1s_2+\epsilon)^\alpha \left( \frac{s_1 + \tau}{s_2+\tau}{\bf u}_1'{\bf u}_1 + \frac{s_2 + \tau}{s_1 + \tau}{\bf u}_2'{\bf u}_2 \right),
\end{equation}
where $\epsilon$ and $\tau$ are set at $10^{-1}$ and $1$ respectively, following \cite{Seo2011face}. The parameter $\alpha$ can be tweaked to boost or suppress the local gradient information depending on the presence of noise. A closed form solution to compute the regularized form of ${\bf C}_{\Omega_i}$ as shown in (\ref{Creg}) is also included in \cite{Seo2011face}.

Finally, the LSK is defined by the following similarity function between the center pixel ${\bf x}_i$ and its surrounding $p\times p$ local neighborhood ${\bf x}_j$, normalized as given below, % $\mathcal{N}({\bf x}_i) = \{{\bf x}_j\}_{j=1}^{p^2}$, given by,
\begin{equation}
\label{LSK}
h_{ij} = \frac{\exp(-\Delta{\bf x}_{ij}' {\bf C}_{\Omega_j} \Delta{\bf x}_{ij})}{\sum_j \exp(-\Delta{\bf x}_{ij}' {\bf C}_{\Omega_j} \Delta{\bf x}_{ij})}, ~~ j = 1, 2, \ldots, p^2,
\end{equation}
where $\Delta {\bf x}_{ij} = \left[ {\bf x}_i - {\bf x}_j \right]'$. The LSK values thus computed at ${\bf x}_i$ are concatenated into a $l = p^2$ dimensional vector ${\bf h}_i$ as follows: ${\bf h}_i = \left[h_{i1}~ h_{i2}~ \ldots h_{il}\right] \in \mathbb{R}^l$. Usually $p\times p$ is considered same in size as that of $\Omega_j$ and is set at $5\times 5$.

Note multiple sources of motivation exist to arrive at the expression of LSK (\ref{LSK}). A detailed treatment is the out of the present scope, but it is worth mentioning that LSK can be motivated and derived from a geodesic interpretation of signal manifold \cite{biswas2016one}, the kernel view of filtering \cite{milanfar2013tour}, and definitely from the idea and definition of structure tensor \cite{peyre2010geodesic}.

Irrespective of all the different sources of motivation and derivation the physical significance explaining the functionality of LSK remains plane and simple: aggressively capturing the locally dominant pattern. Thus it comes as no surprise why LSK features can better retain the overall geometry of the signal manifold in contrast to HOG, illustrated in Fig. \ref{F1}. The comparative visualization in Fig. \ref{F1} confirms that the aggregation of local gradients to estimate principal orientation pattern is able to encode the local geometry exceedingly well as compared to the histogram based statistics of HOG.
\begin{figure}[!t]
\begin{center}
\includegraphics[width=\linewidth]{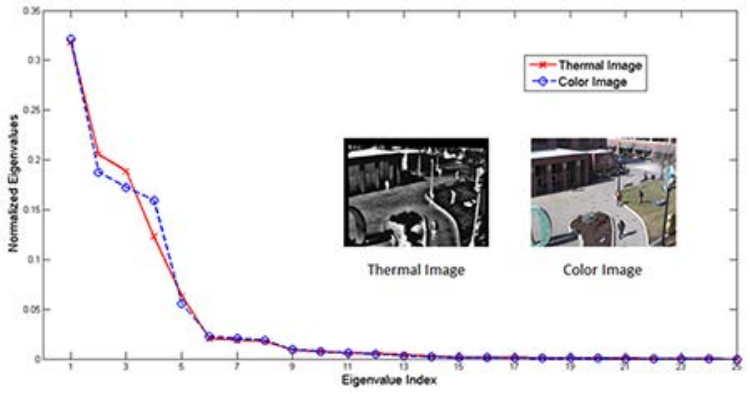}
\end{center}
 \caption{LSK descriptors belong to a low dimensional manifold where 70$\%$ to 80$\%$ of the energy of the eigenvalues is contained in first three or four of them.}
 \label{F2}
\end{figure}
\section{Decorrelation of Local Descriptors}
% If we regard thermal infrared image as ${\bf I}$ as a two diomensional tensor of order $M\times N$, feature computation can be defined as the mapping from a second order tensor to a third order tensor, i.e., $f : \mathbb{R}^{M\times N} \rightarrow \mathbb{R}^{M\times N\times d}$. We accomplish this objective by the following process.
The descriptor vectors ${\bf h}_i$ are stacked together as mode-3 fibers \cite{BaKo06, kolda2009tensor} of a third order descriptor tensor ${\bf H} \in \mathbb{R}^{M\times N\times l}$. Every $i$-th channel of ${\bf H}$, where $i = 1,2, \ldots, l$, encodes some directional property of the image ${\bf I}$. For example, some channels in ${\bf H}$ exhibit vertical structures and some horizontal ones, whereas others capture various oblique directions to a varying degree (Fig. \ref{F1} would give a fair idea).

Besides exhibiting the directional characteristics in them the channels are also observed to be sparse. This behavior is directly reflected in Fig. \ref{F2} where the spectral decomposition of LSK features ${\bf h}_i$ reveals that most of the spectrum energy is stored in the leading few eigenvalues.

In the next step we project the high dimensional tensor ${\bf H}$ onto the principal subspaces along its third mode \cite{kolda2009tensor, lu2011survey}. To be specific, we collect the set of $d$ eigenvectors (computed from LSK descriptors ${\bf h}_i$) as columns of ${\bf V} = \left[ {\bf v}_1 {\bf v}_2 \ldots {\bf v}_d \right] \in \mathbb{R}^{l\times d}$. We choose the number of eigenvalues $d$ in such a way that 80\% of the spectral information is contained in the chosen eigenvalues $\lambda$ as follows: $d = \underset{i} {\mathrm{argmin}} \frac{\lambda_i}{\sum_j \lambda_j} > 80\%$.

Following the \emph{n-mode} product between a higher order tensor and a matrix \cite{lu2011survey, guo2012tensor} we compute the 3-mode product (denoted by $\times_3$) between the descriptor tensor ${\bf H}$ and subspace ${\bf V}$. In other words, the mode-3 ($l$-dimensional) fibers of ${\bf H}$ are projected on the column space of ${\bf V}$. As a result, we obtain the feature tensor ${\bf F} \in \mathbb{R}^{M\times N\times d}$, where $d << l$, given by the following tensor-matrix product:
\begin{equation}
{\bf F} = {\bf H} \times_3 {\bf V}.
\label{E5}
\end{equation}
Doing this has two imminent benefits: one, the projected descriptors are clean, prominent and discriminating (Fig. \ref{F1}), and two, reducing the number of feature channels in ${\bf F}$ has the runtime benefit for fast detection.

\section{Design of Linear Detector with MCS Kernel}
\label{LD}
In the context of one shot object detection, Seo \emph{et al.} \cite{Seo2010training} and Biswas \emph{et al.} \cite{biswas2016one} have shown the effectiveness of Matrix Cosine Similarity (MCS) as a decision rule for computing similarity between two feature tensors (of same size). This measure of image similarity is in fact a generalization of cosine similarity from vector features to matrix/tensor features based on the notion of \emph{Frobenius Inner Product} $\langle \cdot, \cdot \rangle_F$. In principle, suppose ${\bf F}_Q \in \mathbb{R}^{m\times n\times d}$ is a query feature tensor that we try to find in a bigger target tensor ${\bf F}$ in a sliding window fashion. At each position ${\bf x}_i$ of the sliding window over target ${\bf F}$ we compute MCS ($\rho$) as follows:
\begin{equation}
\label{mcs}
\rho({\bf F}_Q, {\bf F}({{\bf x}_i})) = \Biggl\langle \frac{{\bf F}_Q}{\|{\bf F}_Q\|}, \frac{{\bf F}({{\bf x}_i})}{\|{\bf F}({{\bf x}_i})\|} \Biggr\rangle_F,
\end{equation}
where $\|\cdot \|$ denotes \emph{Frobenius norm} for tensors. Higher the value of the MCS at location ${\bf x}_i$ in target image, greater is the likelihood of finding the object there. The normalization allows MCS to focus on phase (or angle) information while also taking care of the signal strength. Besides generalizing cosine similarity, this measure also overcomes the inherent disadvantage of conventional Euclidean distance metric which is sensitive to outliers \cite{fu2008correlation, fu2008image, ma2007discriminant}.

It is important to note that MCS also serves as a valid kernel. To show how, it is quite straight forward to write (\ref{mcs}) in terms of an inner product between two vectors: $\rho({\bf F}_Q, {\bf F}({{\bf x}_i})) = \left( \frac{\mathrm{vec}({\bf F}_Q)}{\|{\bf F}_Q\|}\right)' \left( \frac{\mathrm{vec}({\bf F}({{\bf x}_i}))}{\|{\bf F}({{\bf x}_i})\|} \right)$, where $\mathrm{vec}(\cdot)$ denotes the conventional vectorization operation by stacking the elements of matrix into a long vector. Following this, the proof of MCS being a valid kernel becomes trivial \cite{lampert2009kernel, Scholkopf2001}.

At its core MCS serves as a cross correlation operator between two tensor signals followed by a normalization of signal strength. Closer inspection reveals that such correlation can be performed separately along each feature channel of the third order tensors. The channel correlations can then be combined by summing them up in the next step. To convey the idea, the \emph{Frobenius Inner Product} of (\ref{mcs}) is written below as a multichannel cross correlation,
\begin{align}
& \rho({\bf F}_Q, {\bf F}({{\bf x}_i})) \nonumber \\
&= \sum_{d} \sum_{n} \sum_{m} \frac{{\bf F}_Q(m, n, d)}{\|{\bf F}_Q\|} . \frac{{\bf F}(x_{i1} + m, x_{i2} + n, d)}{\|{\bf F}({\bf x}_i)\|}, \label{mcs2_1} \\
&= \frac{\sum_{d} \sum_{n} \sum_{m} \frac{{\bf F}_Q(m, n, d)}{\|{\bf F}_Q\|} . {\bf F}(x_{i1} + m, x_{i2} + n, d)}{\|{\bf F}({\bf x}_i)\|}. \label{mcs2_2}
\end{align}
The quantity $\|{\bf F}({\bf x}_i)\|$ can be pulled out of the summation as it happens to be the Frobenius norm of the tensor located at ${\bf x}_i$ of the target, and as such, it does not depend on the interaction between ${\bf F}_Q$ and ${\bf F}({{\bf x}_i})$. This form of the kernel as well and the idea expressed above will be used later to facilitate exact acceleration of MCS computation, described in detail in Section \ref{beyond}.
\begin{figure*}[t]
\begin{center}
\includegraphics[width=\linewidth]{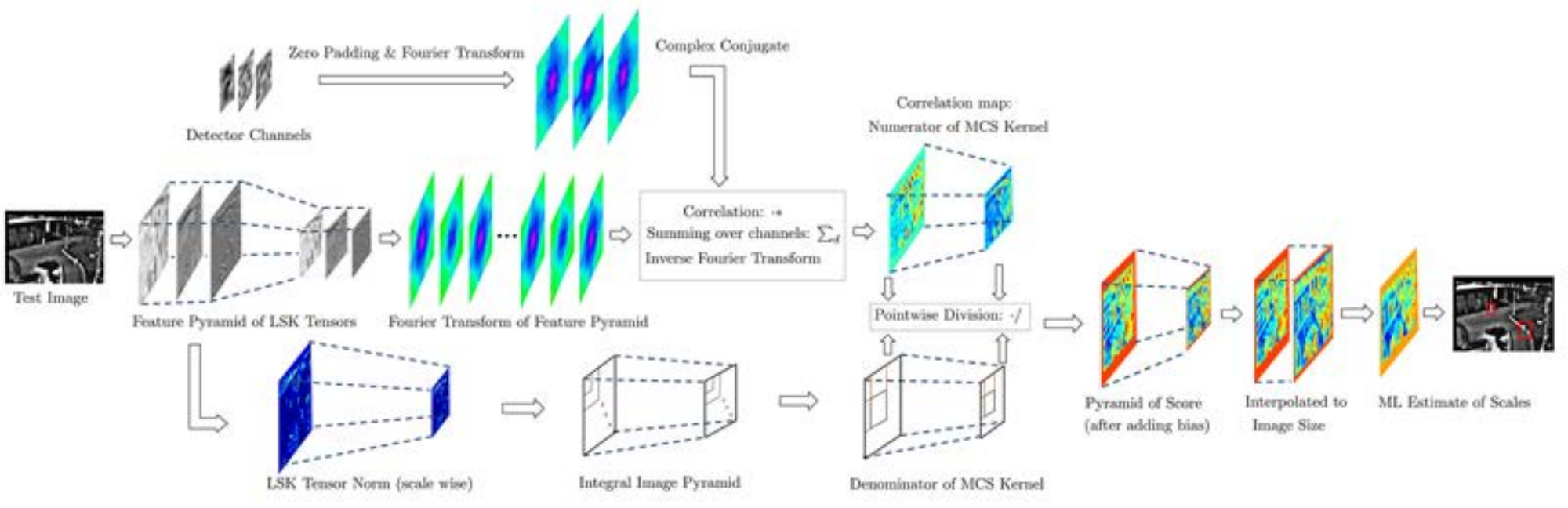}
\end{center}
 \caption{Multiscale detection technique involving construction of feature pyramid, computation of kernel function and maximum likelihood estimate of scale and location of pedestrian in target image}
 \label{F3}
\end{figure*}
\subsection{Linear Support Tensor Machine}
Following the system overview in Section II, we construct our feature set ${\bf F}_i \in \mathbb{R}^{m\times n\times d}$ by first cropping from a full feature tensor (\ref{E5}) according to the ground truth, and next by normalizing with its \emph{Frobenius norm} $\|{\bf F}_i\|$. Paying the polite nod to a slight notational abuse we write the final feature tensor in a slightly overloaded form, ${\bf F}_i \coloneqq \frac{{\bf F}_i}{\|{\bf F}_i\|}$. The learning problem consists of deriving a decision rule based on the set of labeled examples $\mathcal{D} = \{{\bf F}_i, y_i\}_{i = 1 }^N$, where ${\bf F}_i$ has the associated label $y_i \in \{+1, -1\}$, representing one of the two classes, i.e., the pedestrian or the background.

We start our maximum margin formulation by noting that the linear classifier in vector space $\mathbb{R}^d$ is represented by $f({\bf a}; {\bf w}, b) = {\bf a}'{\bf w} + b$. A reasonable way to extend such concept of linear classifier from the vector to the tensor space is the following:
\begin{equation}
\label{lindecrule}
f({\bf F}; {\bf W}, b) = \langle {\bf F}, {\bf W} \rangle_F + b,
\end{equation}
where ${\bf W} \in \mathbb{R}^{m\times n\times d}$ is a third order tensor template that we aim to learn from the annotated examples $\mathcal{D}$. In general, the resulting optimization that follows to learn ${\bf W}$ is given below in its general form:
\begin{equation}
\label{learn}
\underset{W, b} {\mathrm{minimize}~} \frac{1}{2} \|{\bf W}\|^2 + C \sum_{i = 1}^N \mathcal{L} (y_i f({\bf F}_i; {\bf W}, b)),
\end{equation}
where $\mathcal{L}(\cdot)$ denotes the loss over data and $C>0$ is a tradeoff between regularization and constraint violation. Though not critical in the sense that there exists many losses, we have used hinge loss because of its simplicity and wide usage. Henceforth we would assume, $\mathcal{L}(x) = \mathrm{max}(0, 1 - x)$.

Note the MCS kernel can be expressed as an inner product between two vectors. Hence it leads to positive definite real valued kernel with corresponding Reproducing Kernel Hilbert Space (RKHS).  Also, the regularization in (\ref{learn}), in the form of \emph{Frobenius norm} $\|{\bf W}\|$, is a generalization of vector norm to tensor space, and thus can be shown monotonically increasing real valued function. Under such circumstances, the solution $\widetilde{\bf W}$ to (\ref{learn}) can also be written as a linear combination of feature tensors as the direct result of the well known Representer theorem \cite{lampert2009kernel, Scholkopf2001}:
\begin{equation}
\widetilde{\bf W} = \sum_{j = 1}^N y_j\beta_j {\bf F_j}.
\label{repthm}
\end{equation}
When we insert (\ref{repthm}) in (\ref{learn}) the optimization takes place over ($\beta_1, \beta_2, \ldots, \beta_N) \in \mathbb{R}^N$ in the dual domain instead of ${\bf W} \in \mathbb{R}^{(m\cdot n \cdot d)}$ in the primal. Following the optimization, we arrive at the desired classifier function below:
\begin{equation}
\label{hf}
f({\bf F}({\bf x}_i); \beta_1, \beta_2, \ldots, \beta_q, \tilde{b}) = \sum_{j=1}^q y_j \beta_j \rho({\bf F}_j, {\bf F}({\bf x}_i)) + \tilde{b},
\end{equation}
where, ${\bf F}({\bf x}_i)$ according to our past notation denotes the feature tensor at location ${\bf x}_i$ inside ${\bf F}$, and $\tilde{b}$ is the minimizer of (\ref{learn}) with respect to $b$ . In short, $q$ (where $q \leq N$) kernel computations are needed to classify a tensor ${\bf F}({\bf x}_i)$ using the $q$ support tensors. For high dimensional dataset (where $N << m\cdot n\cdot d$) a dual solver for training is preferred because one optimizes less number of parameters: $\beta_1, \beta_2, \ldots, \beta_N$. However, with the increase in dataset size, especially with a large $N$, a primal solver (e.g., stochastic gradient descent \cite{bottou2012stochastic, felzenszwalb2010object}) offers attractive benefit in terms of simplicity and smaller cache size.

While dealing with tensors it is important to note that multilinear algebra provides an inner glimpse of each mode (i.e., dimension) of the tensor --- the roles that they supposedly play (e.g., causal factors like illumination and pose \cite{vasilescu2002multilinear}). The idea here is to apply the linear model ${\bf a}'{\bf w} + b$ but separately in each dimension \cite{maybank2007supervised}. This becomes possible by constraining the template tensor ${\bf W}\in \mathbb{R}^{m\times n\times d}$ to be a sum of $R$ rank-1 tensors --- a direct result of CANDECOMP/PARAFAC (CP) decomposition of higher order tensors. This is given by the following,
\begin{equation}
\label{cp}
{\bf W} = \sum_r^{R} {\bf w}_r^{(1)} \circ {\bf w}_r^{(2)} \circ \ldots \circ {\bf w}_r^{(c)},
\end{equation}
where `$\circ$' represents the tensor outer product \cite{kolda2009tensor, guo2012tensor}. Inserting (\ref{cp}) in (\ref{lindecrule}) results in the following form of the classifier:
\begin{align}
f({\bf F}_i; {\bf W}, b) &= \langle {\bf F}_i, \sum_{r = 1}^R {\bf w}_r^{(1)} \circ {\bf w}_r^{(2)} \circ \ldots \circ {\bf w}_r^{(c)} \rangle_F + b, \label{stl} \\
&= \sum_{r=1}^R \langle {\bf F}_i,  {\bf w}_r^{(1)} \circ {\bf w}_r^{(2)} \circ \ldots \circ {\bf w}_r^{(c)} \rangle_F + b. \label {stl_1}
\end{align}
Once all but ${\bf w}_k$ is fixed, $f({\bf F}; {\bf W}, b)$ becomes the familiar problem of learning the linear classifier ${\bf a}'{\bf w}_k + b$. In general, the approach for learning all ${\bf w}_k$ is to estimate ${\bf w}_k$ from a suitable loss function on $\mathcal{D}$ while treating all other ${\bf w}_i$, where $i\neq k$, constant. The learning algorithm proceeds by repeating this step iteratively for all $k$. In the theory of supervised tensor learning this training methodology is known as alternate projection algorithm \cite{maybank2007supervised}, or more popularly in computer vision literature, as coordinate descent algorithm \cite{pirsiavash2009bilinear}. It is worth pointing out that similar methodology has recently been used in tensor regression to estimate object poses \cite{guo2012tensor}.
\begin{figure}[t]
\begin{center}
\subfigure[]{{}\includegraphics[width=0.24\linewidth]{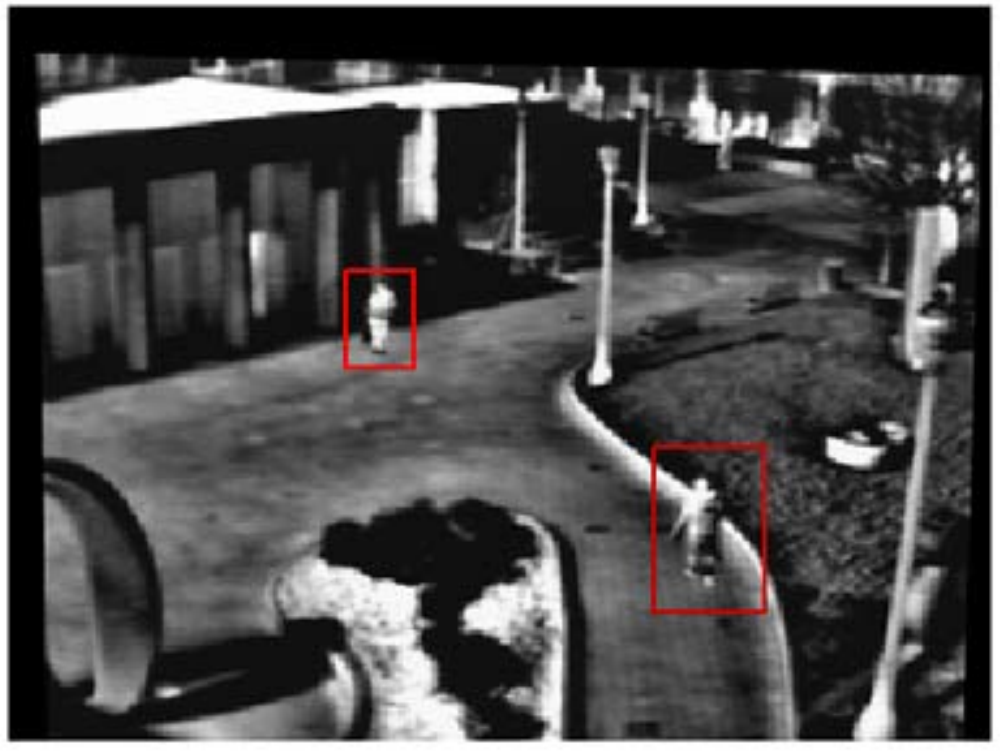}}
\subfigure[]{{}\includegraphics[width=0.24\linewidth]{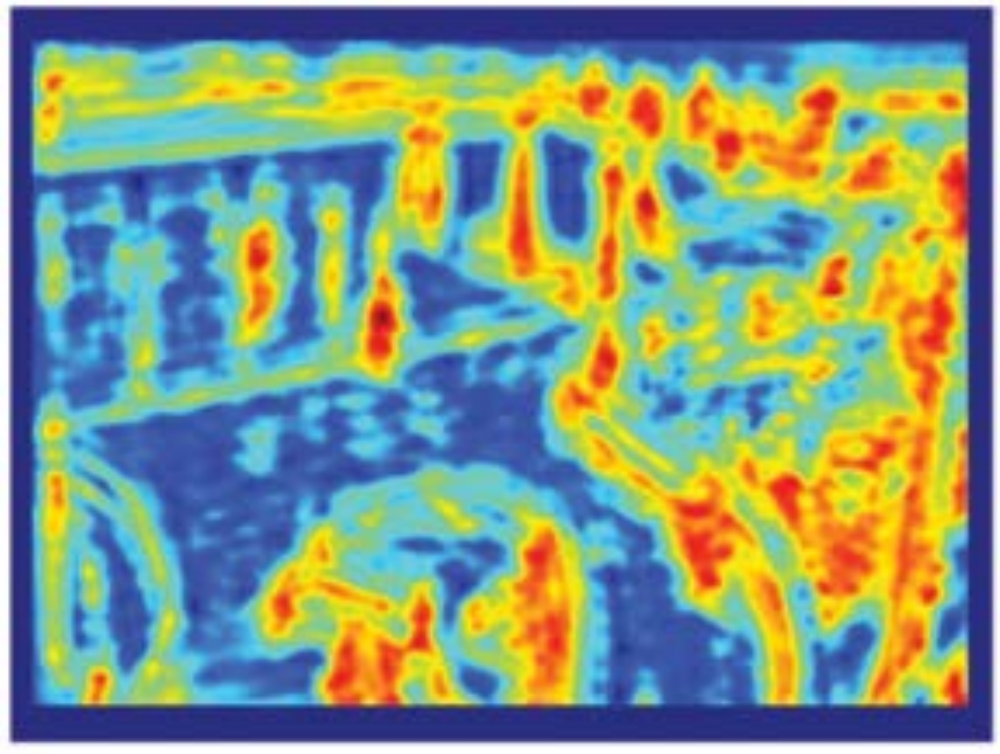}}
\subfigure[]{{}\includegraphics[width=0.24\linewidth]{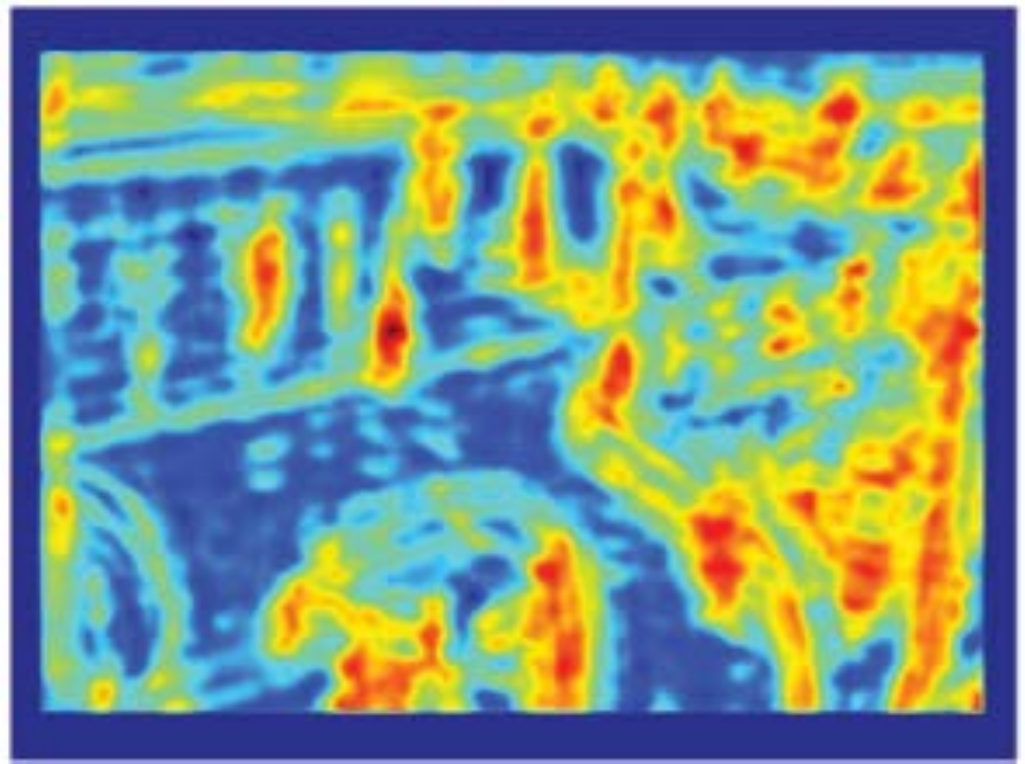}}
\subfigure[]{{}\includegraphics[width=0.24\linewidth]{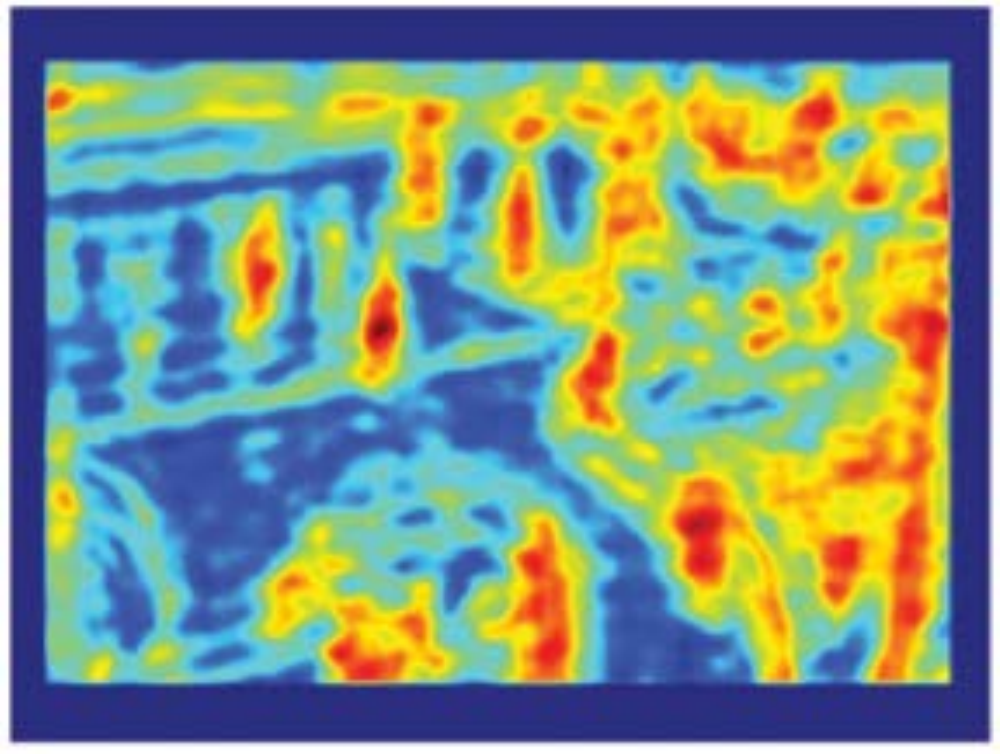}}
\subfigure[]{{}\includegraphics[width=0.24\linewidth]{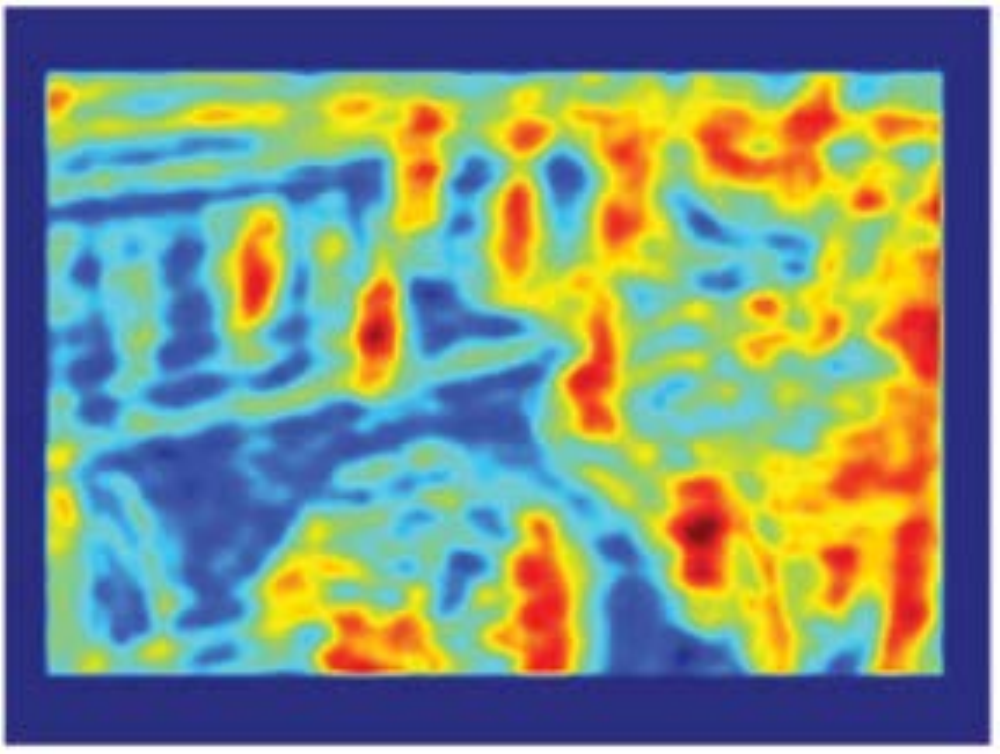}}
\subfigure[]{{}\includegraphics[width=0.24\linewidth]{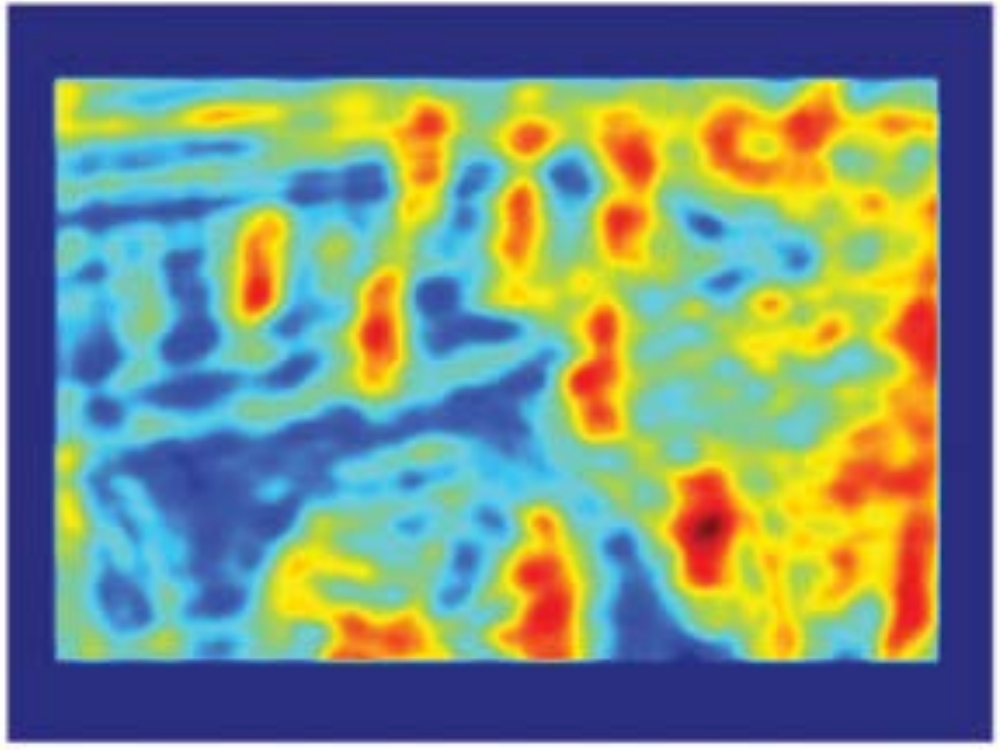}}
\subfigure[]{{}\includegraphics[width=0.24\linewidth]{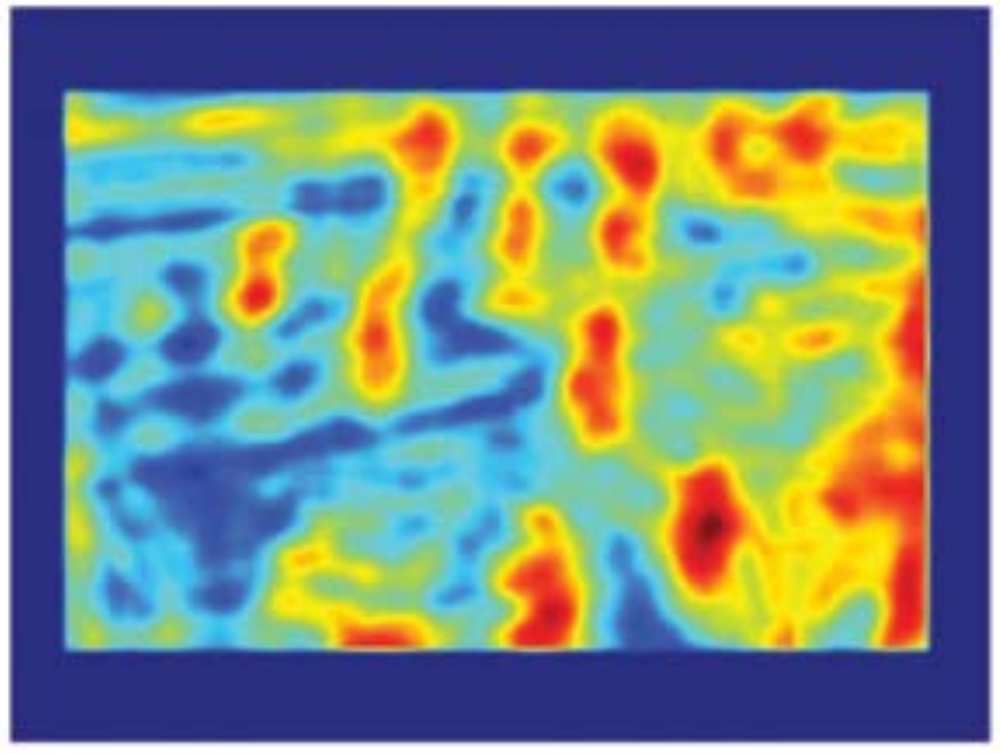}}
\subfigure[]{{}\includegraphics[width=0.24\linewidth]{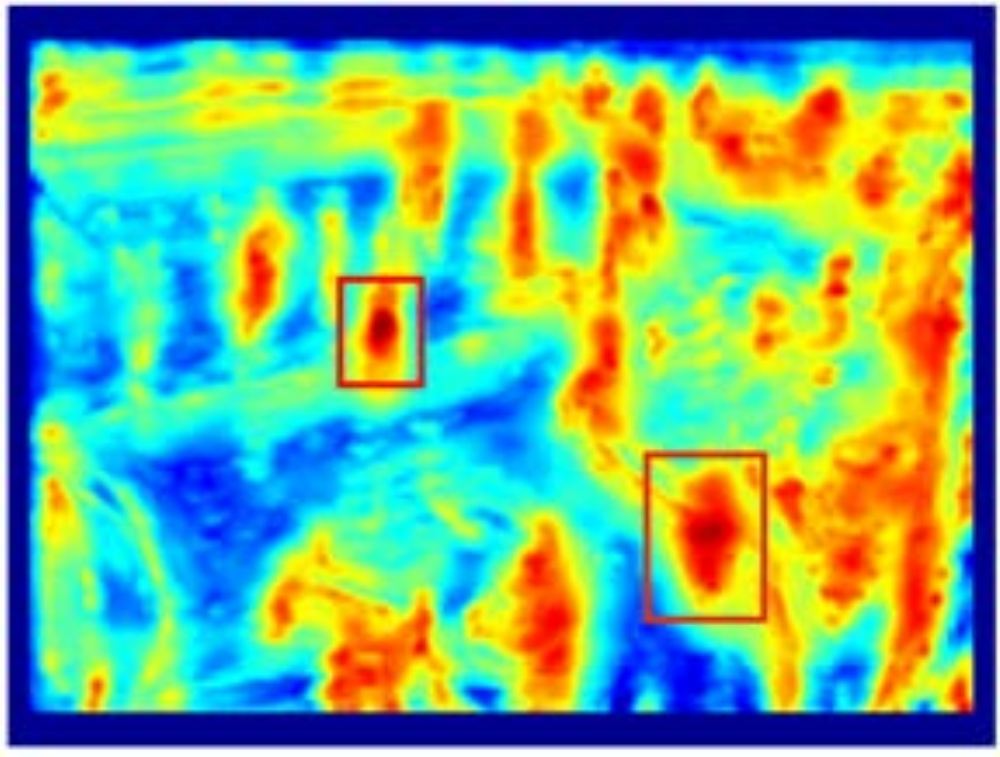}}
\end{center}
   \caption{{\bf Scale Estimation in Multiscale Detection}: Following detection, each scale of features in the feature pyramid yields a score map showing detection score. Note, the boundary region in the score map is getting wider (filled with zeros) with scales because of the decreasing target size in feature pyramid (Fig. \ref{F2}). The individual score maps of various sizes are rescaled with bilinear interpolation to a common size (b)-(g). Lastly, the maximum score at each pixel location is selected from all score maps to obtain the final score map (h), which upon thresholding and non-maximal suppression yields pedestrian location. The score map supplying the maximum score at a particular pixel leads to the scale index associated with the detected bounding box. Blue means low score and dark red to reddish black denote high scores of the classifier function.}
\label{F4}
\end{figure}

It turns out that though multilinear algebra provides a discriminative way to compute different components of ${\bf W}$, the complexity of multilinear support tensor machine is even stricter with total number of parameters being $R (m + n + d)$. Such low complexity no doubt guards the solution from potential overfitting, but also ruins the detector much of its discriminatory power, especially when the number of examples goes high.

In our dataset the number of examples $N$ is comparable to, or even greater than, the complexity of ${\bf W}$. This comes as no surprise because the pedestrians on an average look small in our dataset. At the same time, a reasonably high number of them, along with an equally good number of challenging background examples, motivate us to use a rigid template tensor with decent enough complexity. It appears that number of parameters available in multilinear support tensor machine would be to few to handle the variations present in the dataset. Of course, a trade-off can be achieved by experimenting with an increasing $R$ in the rank-1 approximation (\ref{cp}) of ${\bf W}$. However, it is not clear at this point whether such endeavor is justified in exchange of the much simpler but effective linear model (\ref{repthm})-(\ref{hf}).

The form of classifier (\ref{hf}) provides us further insight in the detector development. To see this, we simplify (\ref{hf}) as follows:
\begin{align}
& f({\bf F}({\bf x}_i); \beta_1, \beta_2, \ldots, \beta_q, \tilde{b}) \nonumber \\
&= \sum_{j=1}^q y_j \beta_j \langle {\bf F}_j, {\bf F}({\bf x}_i)\rangle_F + \tilde{b}, \label{det_1} \\
&= \langle \sum_{j=1}^q y_j \beta_j {\bf F}_j, {\bf F}({\bf x}_i)\rangle_F + \tilde{b}, \label{det_2} \\
&= \langle \widetilde{\bf W}, {\bf F}({\bf x}_i) \rangle_F + \tilde{b} = f({\bf F}({\bf x}_i); \widetilde{\bf W}, \tilde{b}), \label{det_3}
\end{align}
where, $\widetilde{\bf W} = \sum_{j=1}^q y_j\beta_j{\bf F}_j$ as a linear combination of $q$ support tensors forms our detector (with a bias $b$). We are able to write the first step because tensors are normalized and MCS in such case boils down to \emph{Frobenius inner product} (Section \ref{LD}). Second step results by virtue of the linearity of an inner product. In the last step the decision boundary is parameterized in terms of ${\bf W}$ instead of $\alpha_j$.

A few notes follow from our proposed design decisions. First, as also noted by \cite{hao2013linear}, linear support tensor machine makes the training simpler (i.e., one stage) in contrast to multilinear learning, and any off-the-shelf support vector machine solver can handle the optimization problem. Second, even if we are using linear classifiers we deliberately maintain the tensor form of the features. We do not recommend vectorization of tensor features because even if that is permissible for easier training with existing solvers, our detection (prediction) stage would make explicit use of the tensor form for fast and efficient computation. This efficiency resulting from the tensor representation not only aids prediction but also shortens the training time by quickening the hard mining stage.
\section{Exact Acceleration of Tensor Classifier}
\label{beyond}
Searching the detector $\widetilde{\bf W} \in \mathbb{R}^{(m\cdot n \cdot d)}$ in a bigger target tensor ${\bf F} \in \mathbb{R}^{M\times N\times d}$ ($M >> m, N >> n$) by repeated evaluation of the decision rule (\ref{det_3}) is a computationally intensive task. For example, a single channel detector (i.e., $d = 1$) of size $m\times n$ when searched in an $M \times N$ target tensor incurs a computational cost $\mathcal{O}(mnMN)$, and with $d$ feature channels this cost becomes $\mathcal{O}(dmnMN)$. Using the classifier computed in (\ref{det_3}), we ascertain the scores at each pixel position ${\bf x}_i$ of the feature tensor ${\bf F}$ as follows,
\begin{align}
&f({\bf F}({\bf x}_i); \widetilde{\bf W}, \tilde{b}) \nonumber \\
&= \sum_{d} \sum_{n} \sum_{m} \widetilde{\bf W}(m, n, d)\cdot \frac{{\bf F}(x_{i1} + m, x_{i2} + n, d)}{\|{\bf F}({\bf x}_i)\|} + \tilde{b}, \label{drule0} \\
&= \frac{\sum_{d} \sum_{n} \sum_{m} \widetilde{\bf W}(m, n, d).{\bf F}(x_{i1} + m, x_{i2} + n, d)}{\|{\bf F}({\bf x}_i)\|} + \tilde{b}, \label{drule}
\end{align}
The denominator $\|{\bf F}({\bf x}_i)\|$ does not include channel wise interaction with $\widetilde{\bf W}$ and thus can be taken out of the sum. The numerator in (\ref{drule}) involves channel wise cross-correlation (represented by two inner summation) followed by summation across channels (outermost summation). Cross-correlation, especially for a high ratio in sizes between the detector and target tensors, happens very fast in frequency domain. Therefore, to reduce the runtime computation we precompute the channel wise Fourier transforms of the detector tensor $\widetilde{\bf W}$.

During runtime, we perform Fourier transform $\mathcal{F}\{\cdot\}$ of each feature channel ${\bf F}(:, :, i), \forall i = 1, 2, \ldots, d$, perform point-by-point multiplication in frequency domain, the correlation channels thus obtained are summed up right in frequency domain (owing to the linearity of MCS that remains preserved in Fourier transform), and lastly, we invert back the correlation plane in spatial domain by applying inverse Fourier transform $\mathcal{F}^{-1}\{\cdot \}$. The whole process of computing the numerator in (\ref{drule}), for all locations ${\bf x}_i$, is summarized in the following:
\begin{align}
& f({\bf F}({\bf x}_i); \widetilde{\bf W}, \tilde{b}) \nonumber \\
& = \frac{\left[ \mathcal{F}^{-1}\{\sum_d \mathcal{F}\{{\bf F}(:,:, d)\}\cdot \ast \mathcal{F}^\dagger \{ \widetilde{\bf W} (:,:, d)\} \} \right]_{{\bf x}_i}}{\|{\bf F}({\bf x}_i)\|} + \tilde{b}, \label{mcsdft}
\end{align}
where $\mathcal{F}^\dagger \{\cdot \}$ denotes conjugated Fourier transform \footnote{It is worth noting that correlation happens when one of the two Fourier transforms is conjugated, whereas convolution takes place with the product of two Fourier transforms without any conjugation}. The $[\cdot]_{{\bf x}_i}$ in the numerator denotes the correlation score at ${\bf x}_i$ that subsequently gets divided by the normalization factor present in the denominator. Thus computing the numerator of $f({\bf F}({\bf x}_i); \widetilde{\bf W}, b)$ at every pixel location ${\bf x}_i$ in the target tensor takes $\mathcal{O}(dMN\log MN)$ for forward as well as inverse Fourier transform, and $\mathcal{O}(dMN)$ for point by point multiplication as well as for summation. Eventually, we end up with an overall time complexity of $\mathcal{O}(dMN\log MN)$.

The last step in (\ref{mcsdft}) involves normalization by $\|{\bf F}({\bf x}_i)\|$ which could be performed efficiently by computing an integral image of $\sum_d {\bf F}(:,:,d)\cdot \ast {\bf F}(:,:,d)$ involving a time complexity $\mathcal{O}(dMN)$. The retrieval of the normalization value eventually follows from the square root in constant time per window. In summary, the computation cost of the classifier is dominated in the numerator by $\mathcal{O}(dMN\log MN)$, and by $\mathcal{O}(dMN)$ in the denominator, leading to an overall complexity of $\mathcal{O}(dMN\log MN)$. This is reasonably less in contrast to the brute force time complexity of sliding window detection: $\mathcal{O}(dmnMN)$. The fact that the cost no longer relies on the rigid detector's template size results in a substantial gain in efficiency. The complete methodology for evaluating the proposed MCS is illustrated in details in Fig. \ref{F3}.

Such idea of accelerating the detection process by a multichannel implementation of Fourier transform has found recent application in \cite{Fleuret2012exact}, as well as in \cite{biswas2016one}. In \cite{Fleuret2012exact}, the authors did not have to deal with the normalization factor that is present in MCS. Biswas et. al, \cite{biswas2016one} extended their technique to accelerate the MCS computation. In this paper, we show how the fast kernel computation can further be extended to efficient pedestrian detection following a tensor based maximum margin learning setup. The proposed acceleration of decision rule does not involve any approximation, hence, it remains an exact version with a much shorter detection time.
\begin{figure}[t]
\begin{center}
   \includegraphics[width=1.0\linewidth]{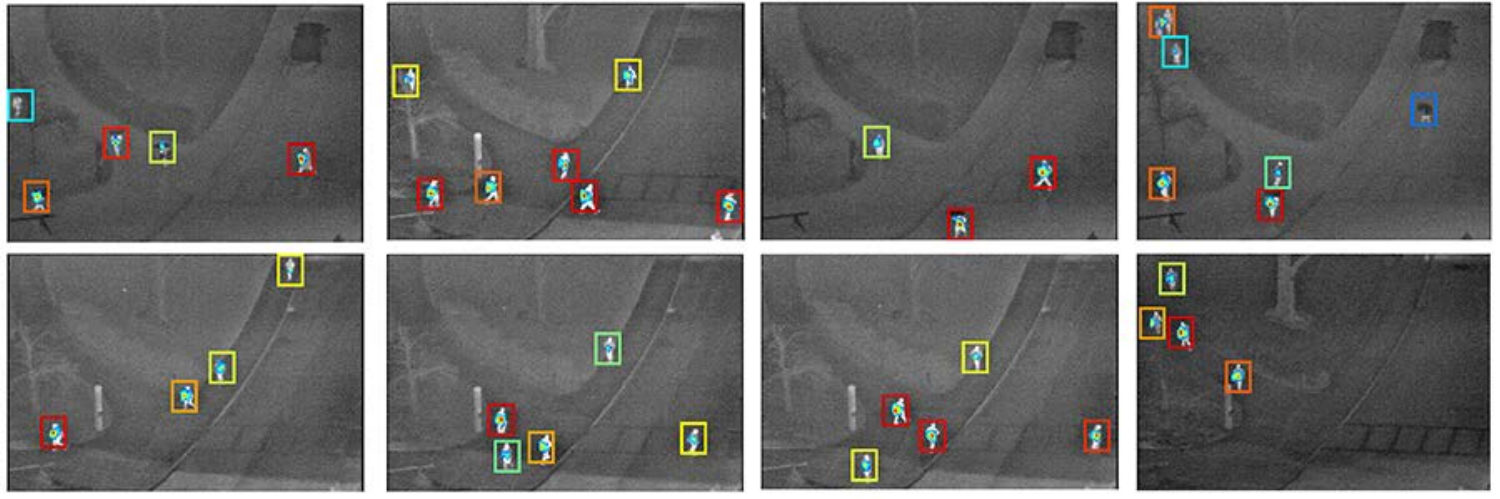}
\end{center}
   \caption{Detection results of our proposed methodology are shown in this figure. The detection scores above the threshold are embedded inside the displayed bounding box. The convention of color map is maintained, i.e., a red bounding box indicates highest confidence and blue bounding box lowest confidence.}
\label{F5}
\end{figure}
\section{Multiscale Detection Methodology}
There are two approaches generally available for multiscale search of objects. One approach is to scale up the rigid detector and search for maximum scoring region. Though attractive because target image undergoes minimum transformation during runtime, from a purely theoretical standpoint this scaling up of a rigid detector can have the uncanny effect of introducing artifacts in bias $b$ while computing $f(F; \widetilde{\bf W}, \tilde{b})$ at every location ${\bf x}_i$. It is not immediately obvious how and to what extent such issues will manifest in the present methodology, and in case they do, what could be probable way out to mitigate such limitation. Hence, we have followed the second approach that involves target rescaling. To be specific, we computed features from the given image and resorted to feature scaling over the desired range of scales. In other words, we have constructed a feature pyramid of decreasing image size as described in Fig. \ref{F3}.

We describe next how we infer the pedestrian's location and the size in the test image. It is important to note that for each scale we essentially obtain a score map as a result of detection (Fig. \ref{F4}). Each pixel intensity in a particular score map represents the value of scoring function of the proposed linear classifier at that particular scale. We rescale the likelihood maps of all scales to bring them to a common size (the largest scale in our case) before selecting the maximum score at each pixel to best estimate the scale that is producing the maximum detection score. The final score map thus obtained is thresholded (usually at zero) following a non-maximum suppression step to output the pedestrian's location. The maximum scale associated with the pedestrian's location provides the size of the bounding box we need.
\begin{figure*}
\begin{center}
\subfigure[]{{}\includegraphics[width=0.3\linewidth]{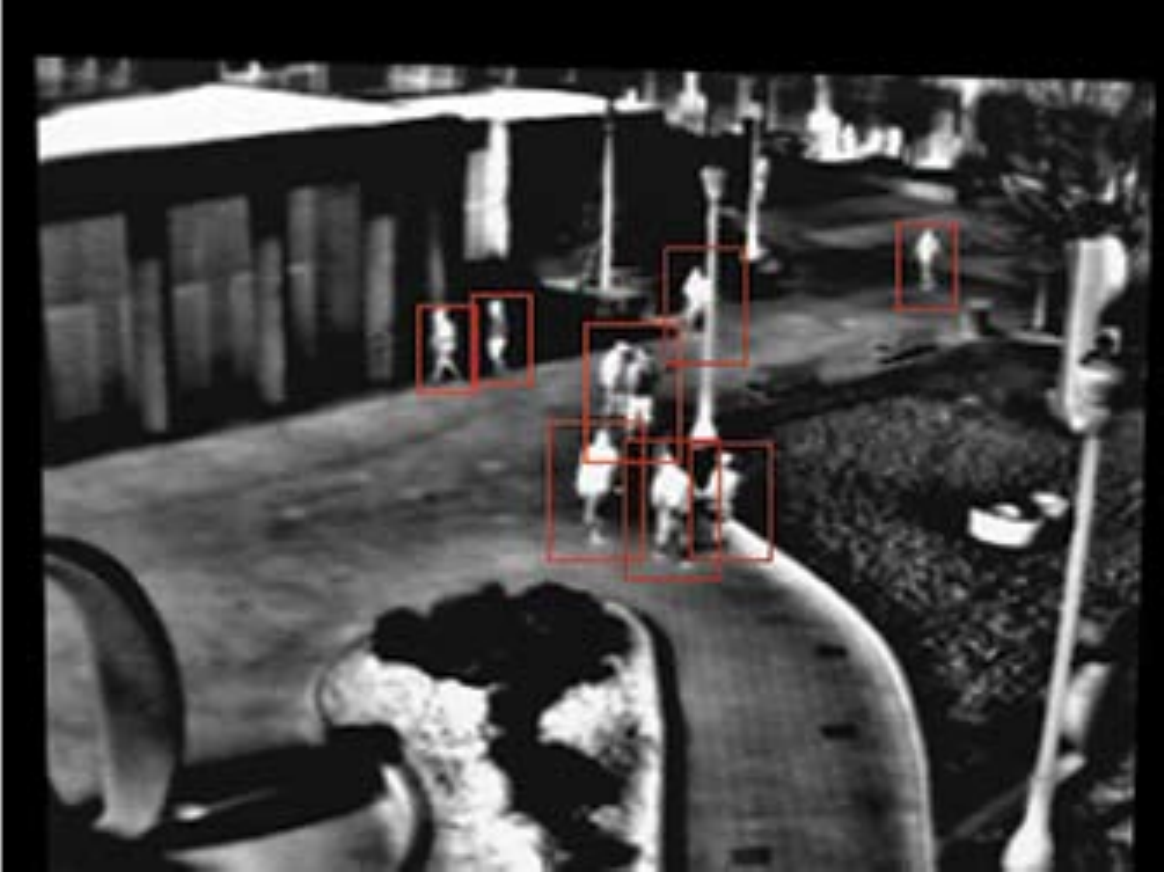}}
\subfigure[]{{}\includegraphics[width=0.3\linewidth]{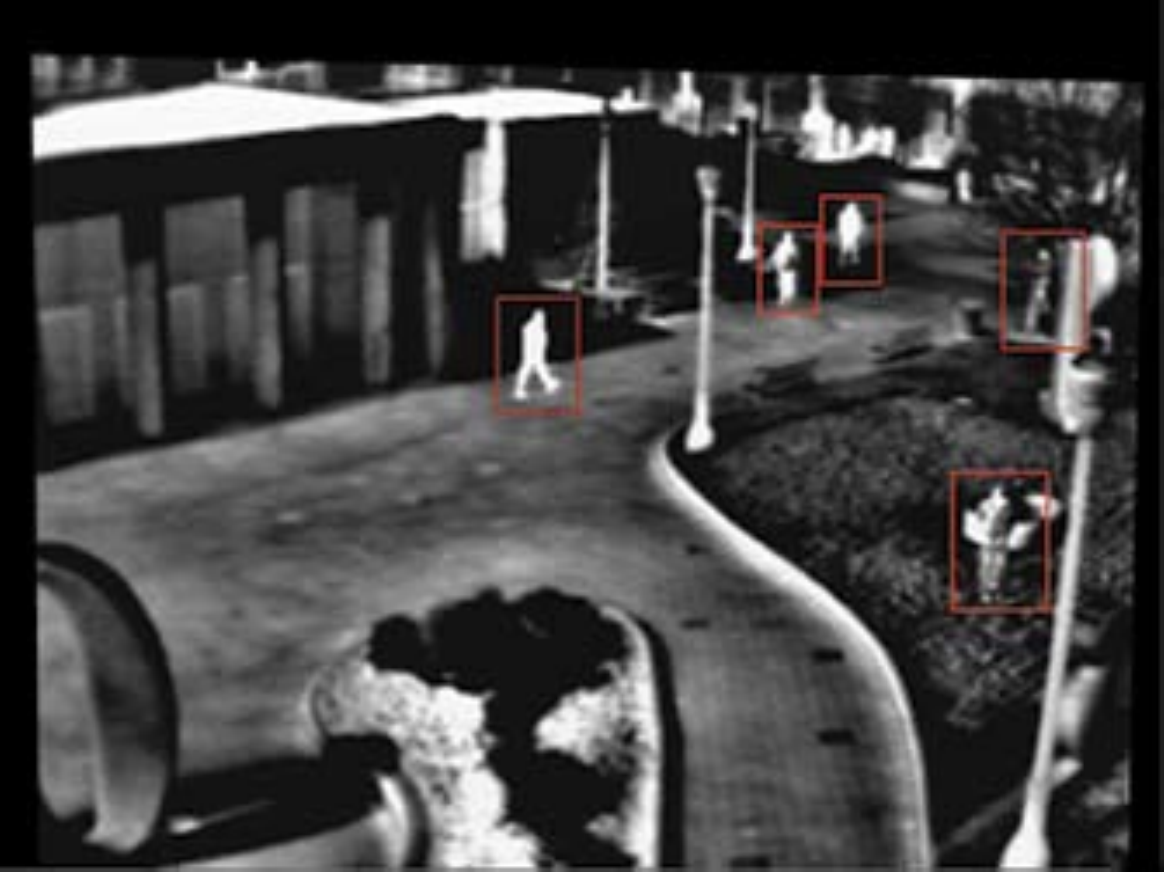}}
\subfigure[]{{}\includegraphics[width=0.3\linewidth]{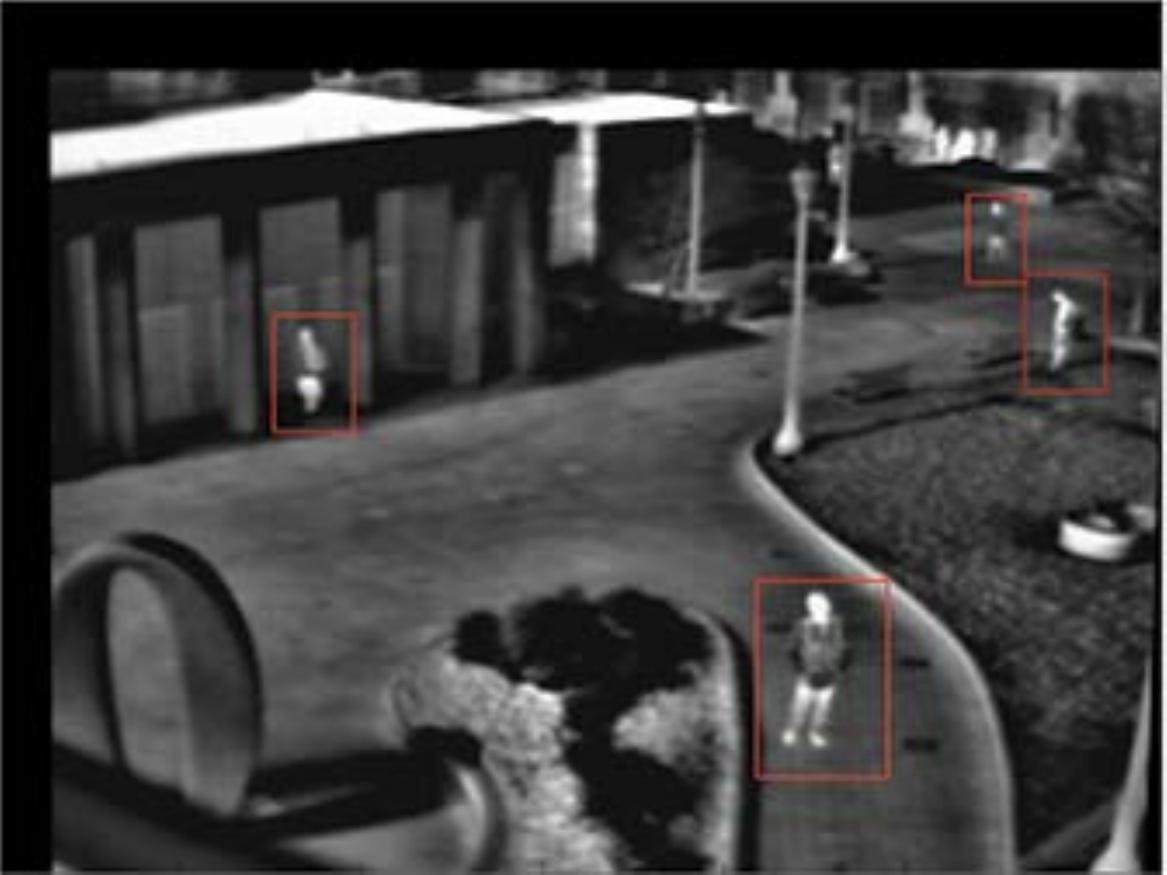}}
\\
\subfigure[]{{}\includegraphics[width=0.3\linewidth]{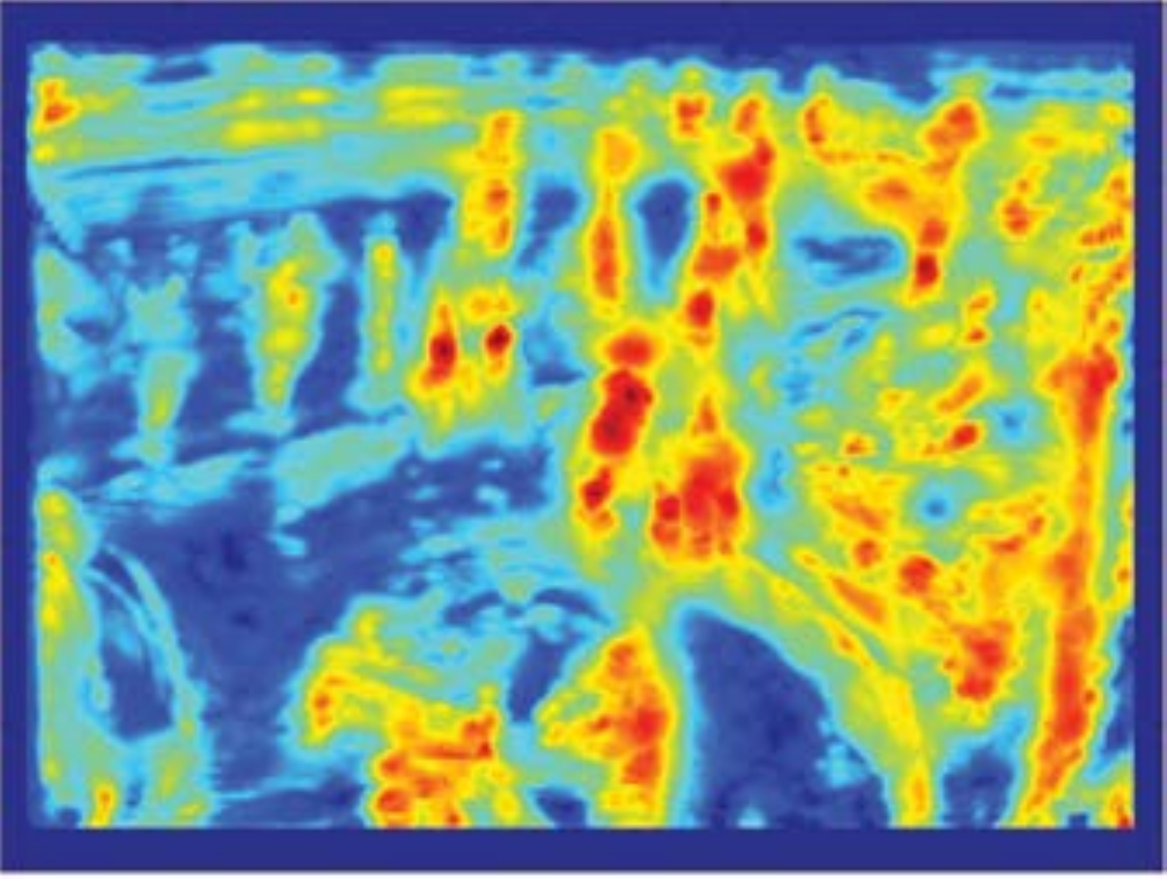}}
\subfigure[]{{}\includegraphics[width=0.3\linewidth]{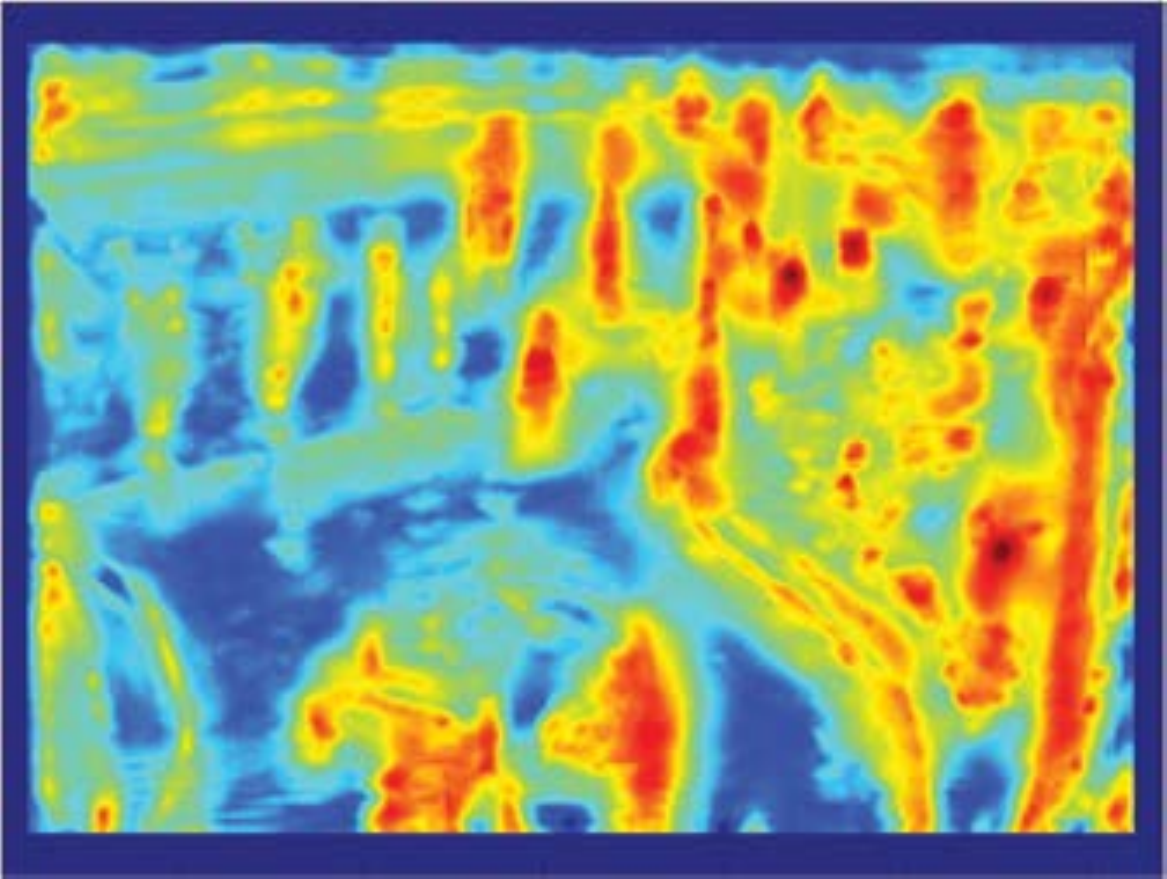}}
\subfigure[]{{}\includegraphics[width=0.3\linewidth]{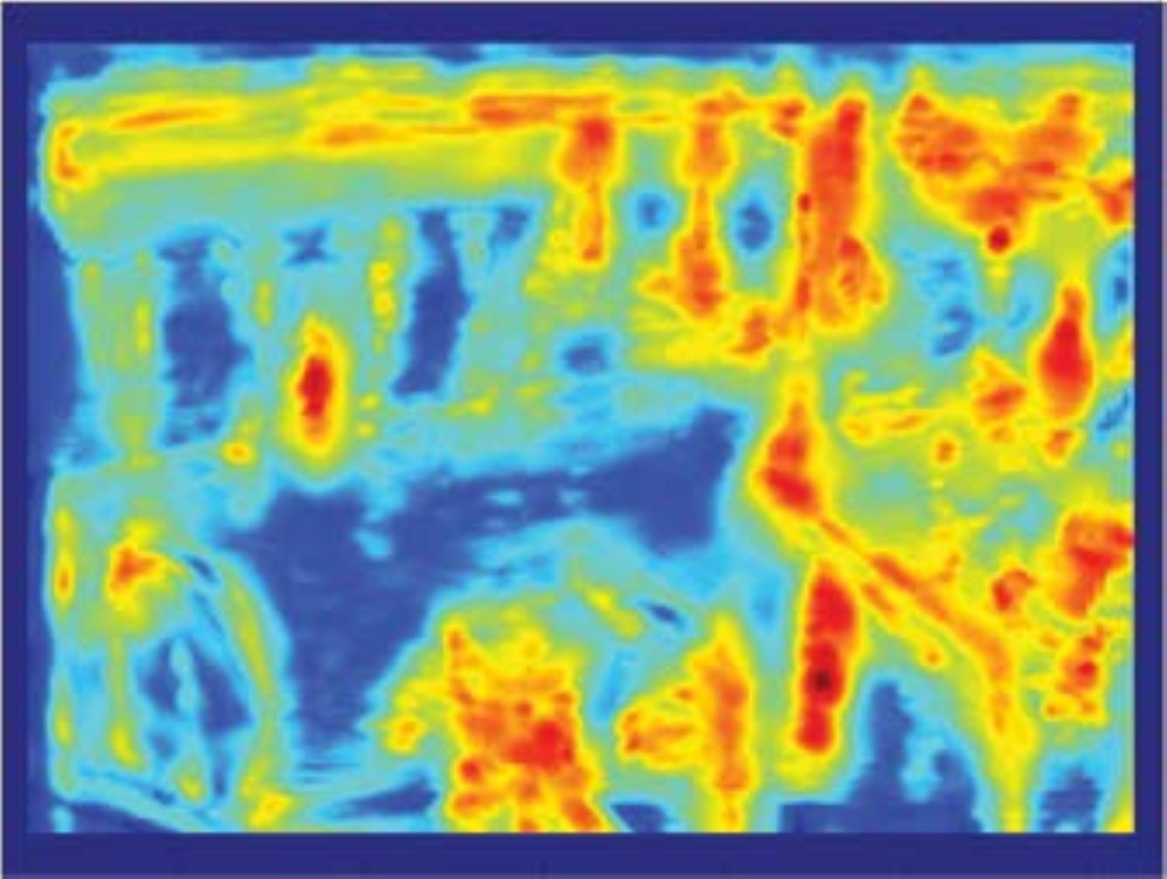}}
\end{center}
   \caption{Top row shows multiscale detection on three frames from OSU-CT dataset. The scale best estimated is shown with the appropriate sized bounding box centered at the predicted location. The bottom row heat maps illustrate corresponding decision scores (maximum likelihood estimate across all six scales) obtained from classifier. The blue regions show less confidence and the red to reddish black shows high to very high confidence in detecting pedestrians. The proposed detector faces difficulty in detecting partially occluded people in absence of any tracking information and/or background model.}
\label{F6B}
\end{figure*}
\section{Experiments and Results}
Though infrared images often exhibit visual cues that remain absent in the visible spectrum, the clarity and usefulness of such information may be limited by several other extraneous parameters like sensor noise, temperature of objects, weather conditions, indoor and outdoor environments. The extent to which computer vision tasks like visual recognition is influenced by those external factors requires a careful study of large-scale, well annotated infrared datasets that are not still as many, and the size of such datasets, where available, is relatively small \cite{terravic, davis2005two}. The ready availability of similar useful resources have facilitated steady improvement in the performance of pedestrian detection in natural images (CalTech \cite{dollar2012pedestrian}, INRIA pedestrians \cite{dalal2005histograms}) over the last few years \cite{benenson2014ten}.

To mitigate this shortcoming new and large-scale thermal image datasets have been developed recently, e.g., LSI \cite{LSI} and KAIST \cite{hwang2015multispectral} and BU-TIV \cite{wu2014thermal}. The decent image quality of LSI and variable heights of the pedestrians offer a good range of difficulty levels to develop pedestrian detectors. KAIST multispectral images, also captured in a  real life setting, shows rapid degradation of image quality in thermal channel with increasing distance, and it gets quite difficult to distinguish distant pedestrians from background with the infrared channel alone (KAIST dataset comes with RGB color channels too). The BU-TIV dataset has relatively high resolution images of human beings for a wide range of visual recognition tasks like detection, single view and multi-view tracking. The objects for the detection task in the BU-TIV not only include pedestrians but also other classes like cars and bikes on a crowded street.

In this work we focus our attention on detecting pedestrians in three baseline datasets: OSU Thermal (OSU-T) \cite{davis2005two}, OSU Color Thermal (OSU-CT) \cite{davis2007background}, and LSI \cite{LSI}. We restrict our study to thermal channels only. Color channels when available are ignored. The baseline dataset OSU-T contains pedestrians in still images. The other two datasets namely OSU-CT and LSI both are infrared video datasets.

In the experimental setup we have first computed the three dimensional LSK descriptors where third dimension denotes the number of descriptor channels. Such number is always 25 since we have considered 5 $\times$ 5 neighborhood around the central pixel in (\ref{LSK}). The number of eigenvectors used for feature computation is typically three unless mentioned otherwise. We have used LIBSVM \cite{CC01a} solver to solve the max-margin optimization task. The choice of a solver is not critical, and the proposed methodology is general enough to solve with any quadratic solver, e.g., \cite{shalev2013stochastic} and \cite{shalev2011pegasos} which are available in VLFeat library \cite{vedaldi08vlfeat}.

For describing the evaluation process we have followed the 50$\%$ intersection-over-union PASCAL criterion \cite{everingham2010pascal}) between the detected bounding box and the supplied ground truth, to determine correct detection and missed detection (or false negative). In particular, we define the miss rate by $FN/(TP+FN)$, where FN represents the total number of false negatives, and TP the total number of true positives. The total number of false positives or FP is normalized by the number of images in the test set leading to FPPI or false positives per image. Following the evaluation technique established for detecting pedestrians in the visible spectrum \cite{dollar2012pedestrian}, we report here the miss rate versus FPPI graph as a measure of detector performance. This is in contrast to earlier notion of false positive per window (FPPW) as used to evaluate the pedestrian detector \cite{dalal2005histograms, hotspot14}. Miss rate versus FPPI is also in contrast to the precision-recall curves that are more traditionally followed in other areas of object detections \cite{everingham2010pascal}. The present evaluation criterion is motivated by the applications like autonomous driving where it is often the norm to fix the upper ceiling at acceptable FPPI rate independent of the number of pedestrians in the image.

{\bf OSU Thermal Database (OSU-T)}: OSU thermal images \cite{davis2005two} come from 10 sequences with a total number of 284 images all of which are 8-bit. This dataset is not a video sequence because the images are captured in a non uniform fashion with a sampling rate less than 30Hz. The image size is 360 $\times$ 240 pixels. In total, the dataset has 984 pedestrians across all 10 sequences.

We have followed a $K$-fold cross validation technique for the evaluation by holding out each of the 10 sequences for the test, and use the rest of the sequences for training. The detection results of each held-out sequence are later combined in a big text file, and analyzed, to compute the overall miss rate and FPPI \cite{dollar2012pedestrian} for the full dataset. It is worth noting that in the wide area surveillance, like satellite image analysis, it is pretty common to attribute less emphasis to scale. Objects at a distance do not appear to vary widely in sizes. Indeed, with a \emph{single scale rigid detector} we have achieved reasonably good performance as shown in the Fig. \ref{F5}. Note that the ground truth supplied varies in height and width across pedestrians. However, we have extracted a constant height $\times$ width bounding box (36 $\times$ 28 to be specific) around the center of the given ground truth rectangle for each pedestrian. With the change in weather condition the appearance (illumination in particular) of the background varies significantly. We built an initial detector with a subset of the pedestrian and background tensor features, and collected hard negative examples (like \cite{dalal2005histograms, felzenszwalb2010object}) by trying to detect pedestrians with the initial detector. The $\alpha$ is set to 0.4 in this experiment following the feature extraction setup of \cite{biswas2016one}.
\begin{figure}
\begin{center}
   \includegraphics[width=1.0\linewidth]{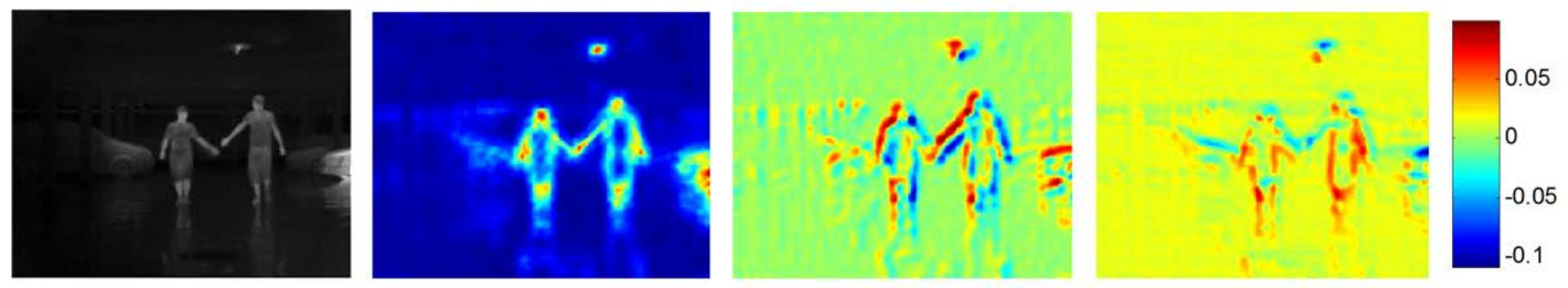}
\end{center}
   \caption{The thermal image on far left is shown in three LSK feature channels on right. Note how the first channel shows signal strength around body silhouette, whereas second channel tends to highlight horizontal to oblique structures. The third channel mostly models the vertical to near vertical structures.}
\label{F7A}
\end{figure}

{\bf OSU Color Thermal Database (OSU-CT)}: OSU Color Thermal dataset \cite{davis2007background} has a total of 17088 images (8-bit thermal and 24-bit color) and is a video sequence dataset. This dataset has a total of six sequences with each three containing scenes of same location. Each of the six sequences has thermal as well as color channels. Since this dataset does not have a ground truth available in the public domain, we have annotated the full data set (using tools \cite{dollar2012pedestrian, dollarCVPR09peds, PMT}) for the evaluation of the proposed detector. The annotation task gets challenging because the pedestrians at a distance often get occluded by physical structures (e.g., poles and tree branches), or by other pedestrians. In such cases, we have either ignored the heavily occluded pedestrian, or approximately sized the bounding box around a partially occluded person to the best guess possible.

The six thermal sequences have a total number of 8544 images. From a purely pedestrian detection perspective, the last three of the six sequences are somewhat irrelevant because a large number of frames have only one or two pedestrians, and sometimes none. We have experimented with the first three sequences (containing 3355 images in total) that are extremely challenging involving the presence of many pedestrians, heavy occlusions, and low-resolution. Following the sampling procedure of CalTech pedestrian dataset \cite{dollar2012pedestrian}, we have uniformly sampled the frames (every 10-th frame) from each of the three video sequences to include in our experiment. In total, we have 1534 pedestrians coming from all the three sequences. Similar to OSU-T we have followed a 3-fold cross validation to complete our evaluation process.

In this dataset, the pedestrians are not as far as those in OSU-T from the camera position. As a consequence, we employ a multi-scale detection strategy to meet our goal. The full dataset contains pedestrians with heights ranging from 14 to 60 pixels. However, we ignore pedestrians which appear too small (less than 20 pixels in height) when they are too far from the camera. We have learnt a rigid detector of size 30 $\times$ 20, and searched it in the target images over the following six scales: 1.30, 1.00, 0.81, 0.68, 0.59, and 0.52. We increase $\alpha$ value to 0.75 to boost the weak singal. An initial detector is computed first to collect hard negative examples from the background of this dataset. We have also applied the initial detector on negative images of LSI to pick hard negatives from this dataset. The final detector is learnt with positive examples from OSU-CT, and hard negative examples from the background of OSU-CT as well as LSI images.

The score maps resulting from the multi-scale detection are shown in Fig. \ref{F6B} in the form of heat color map. Blue denotes very low confidence, whereas dark red to red-black denotes high to very high confidence in predicting a pedestrian. The heavy occlusion, low-resolution and vertical structures in the background make the detection task quite challenging.
\begin{figure*}
\begin{center}
\subfigure[]{{}\includegraphics[width=0.24\linewidth]{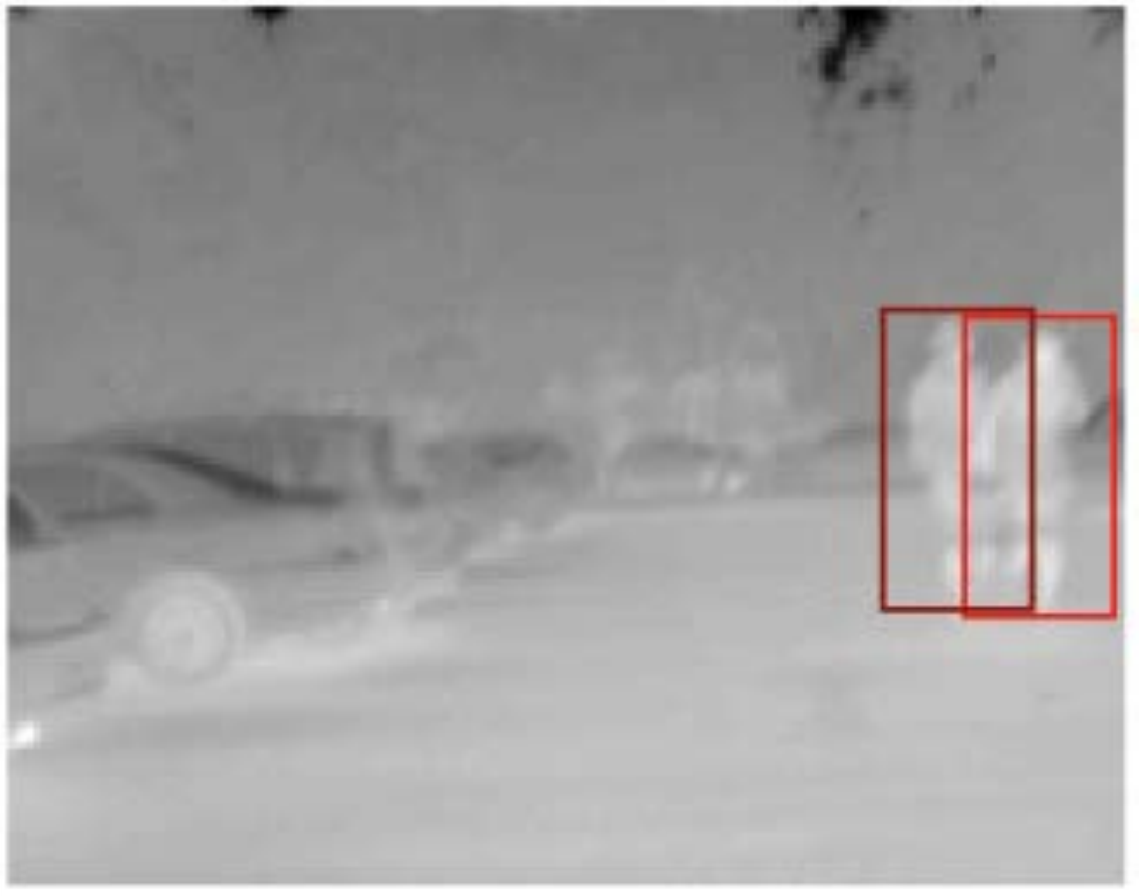}}
\subfigure[]{{}\includegraphics[width=0.24\linewidth]{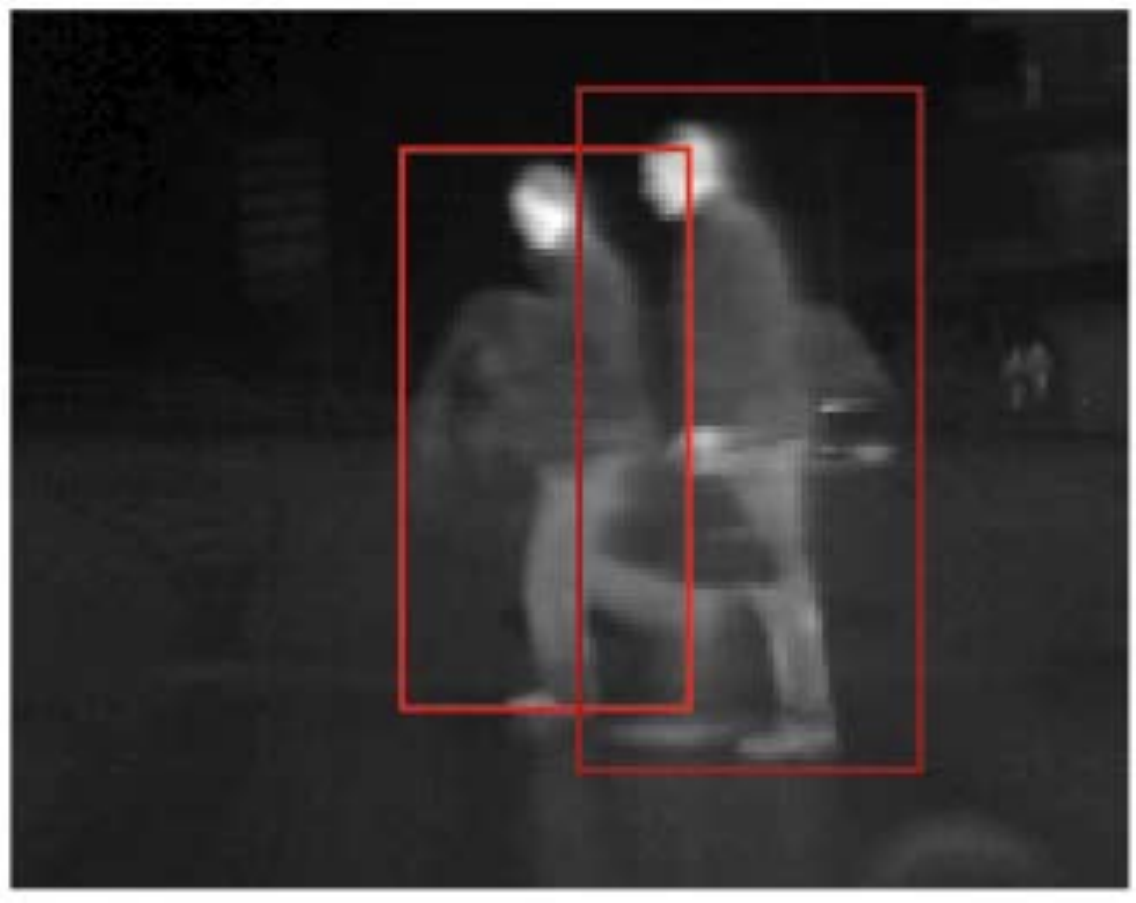}}
\subfigure[]{{}\includegraphics[width=0.24\linewidth]{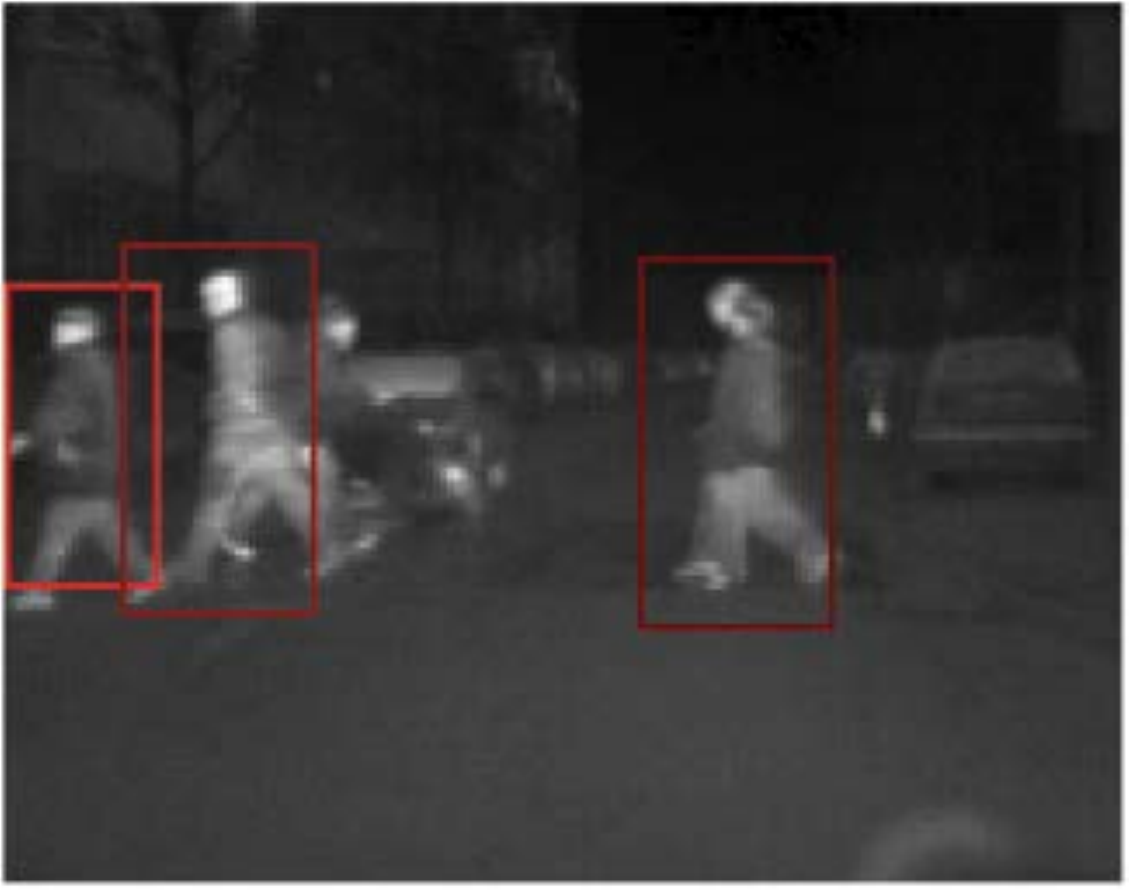}}
\subfigure[]{{}\includegraphics[width=0.24\linewidth]{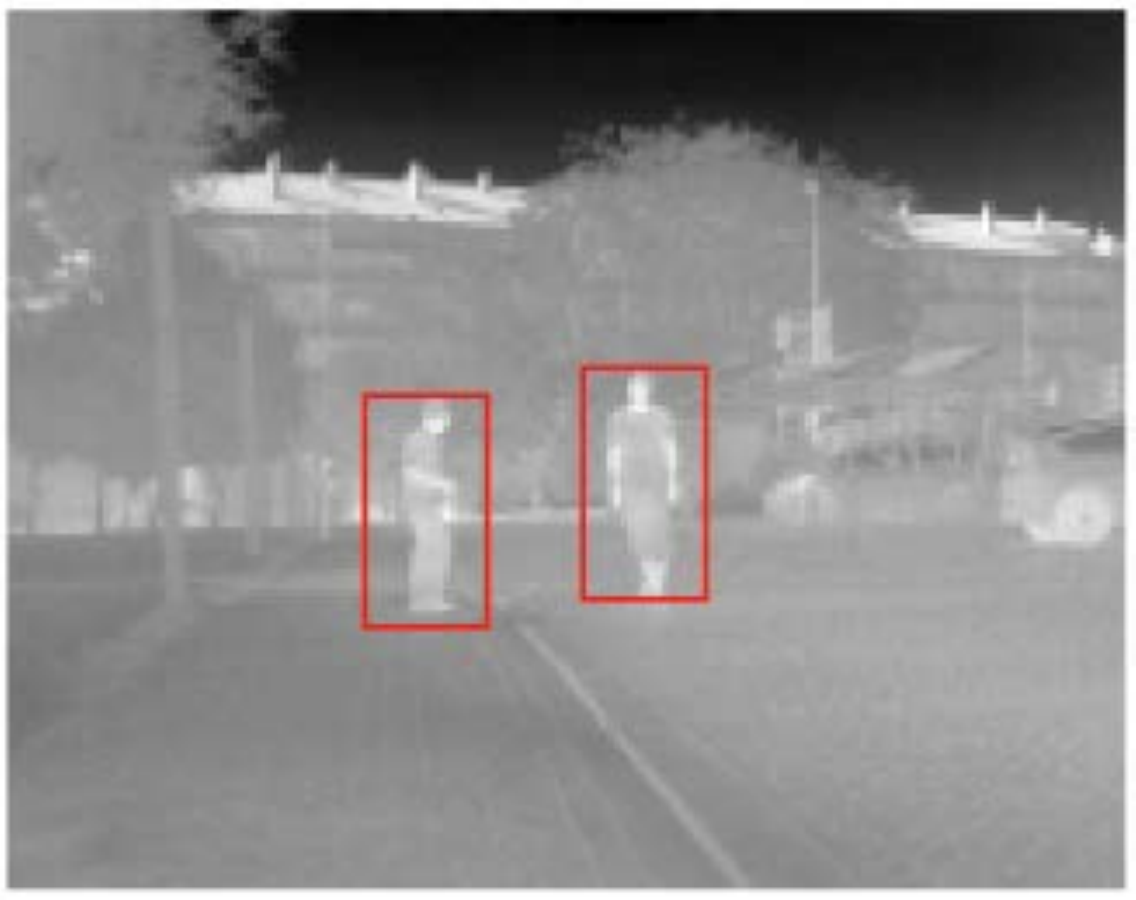}}
\\
\subfigure[]{{}\includegraphics[width=0.24\linewidth]{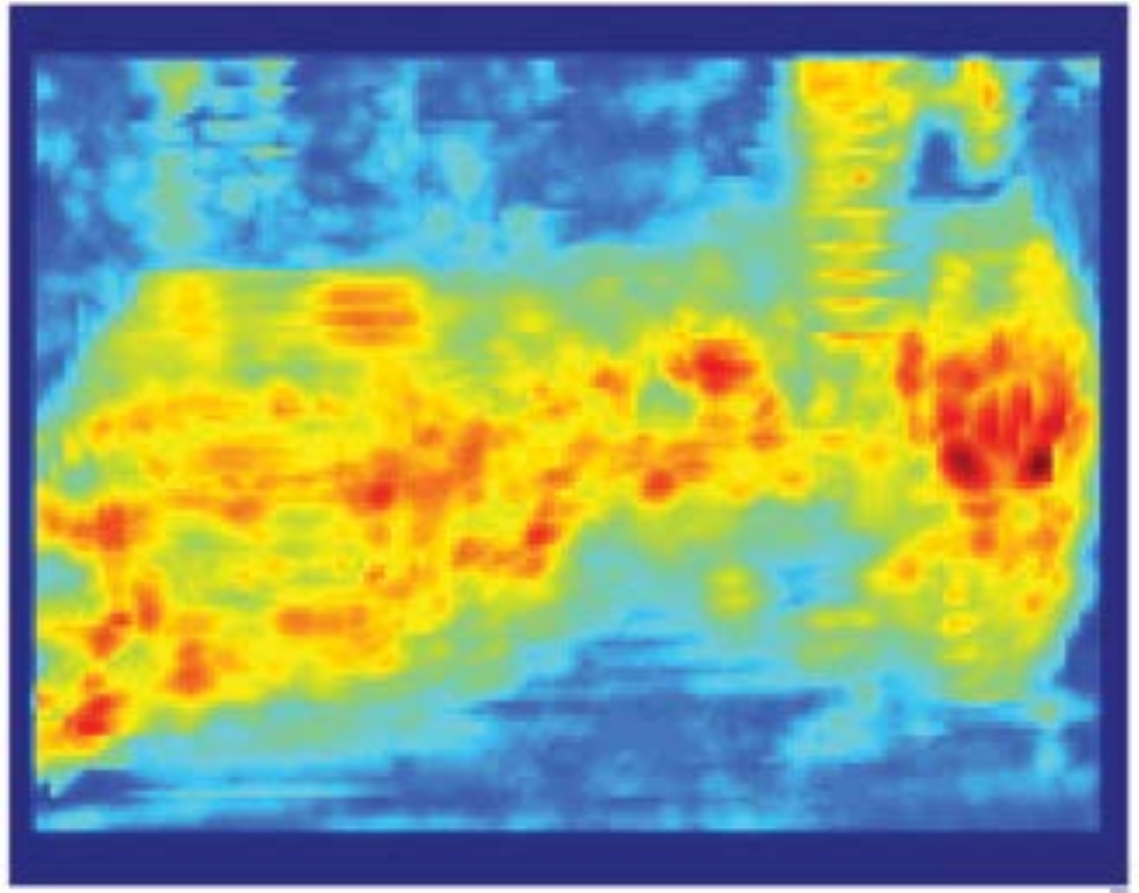}}
\subfigure[]{{}\includegraphics[width=0.24\linewidth]{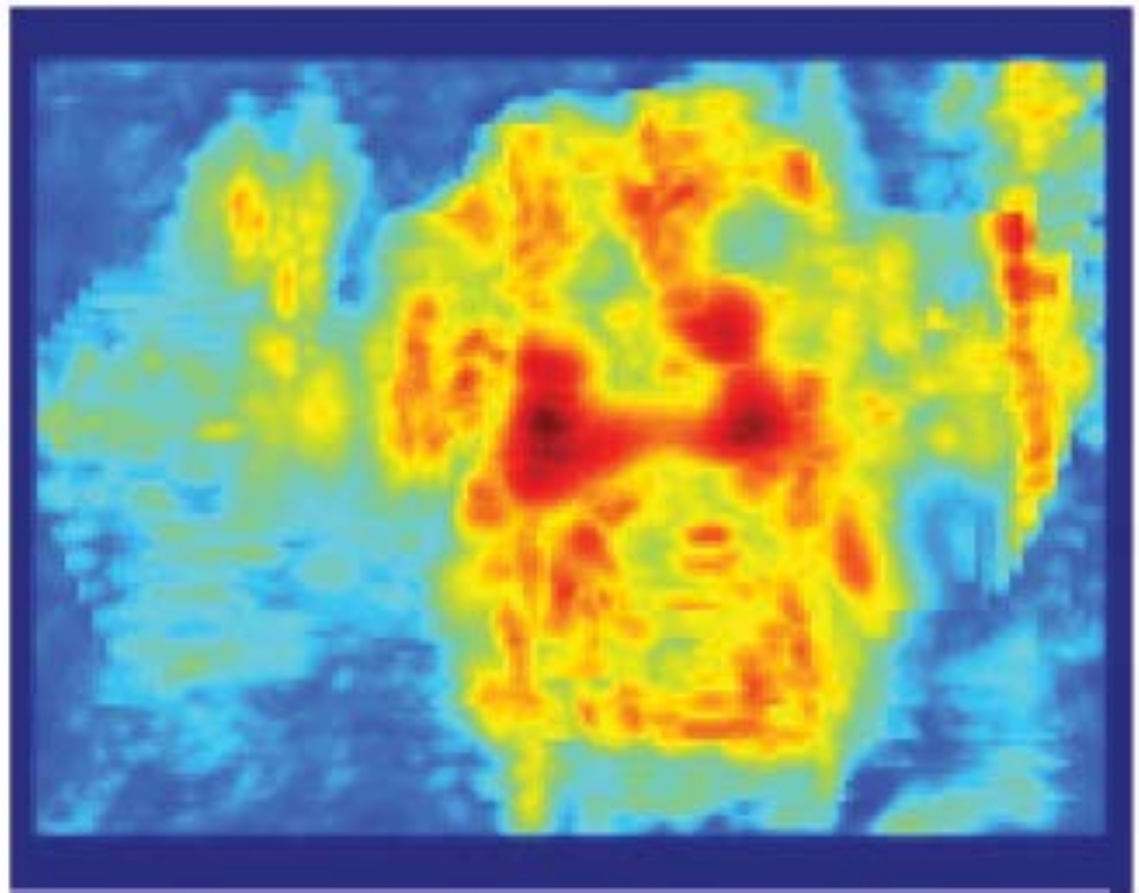}}
\subfigure[]{{}\includegraphics[width=0.24\linewidth]{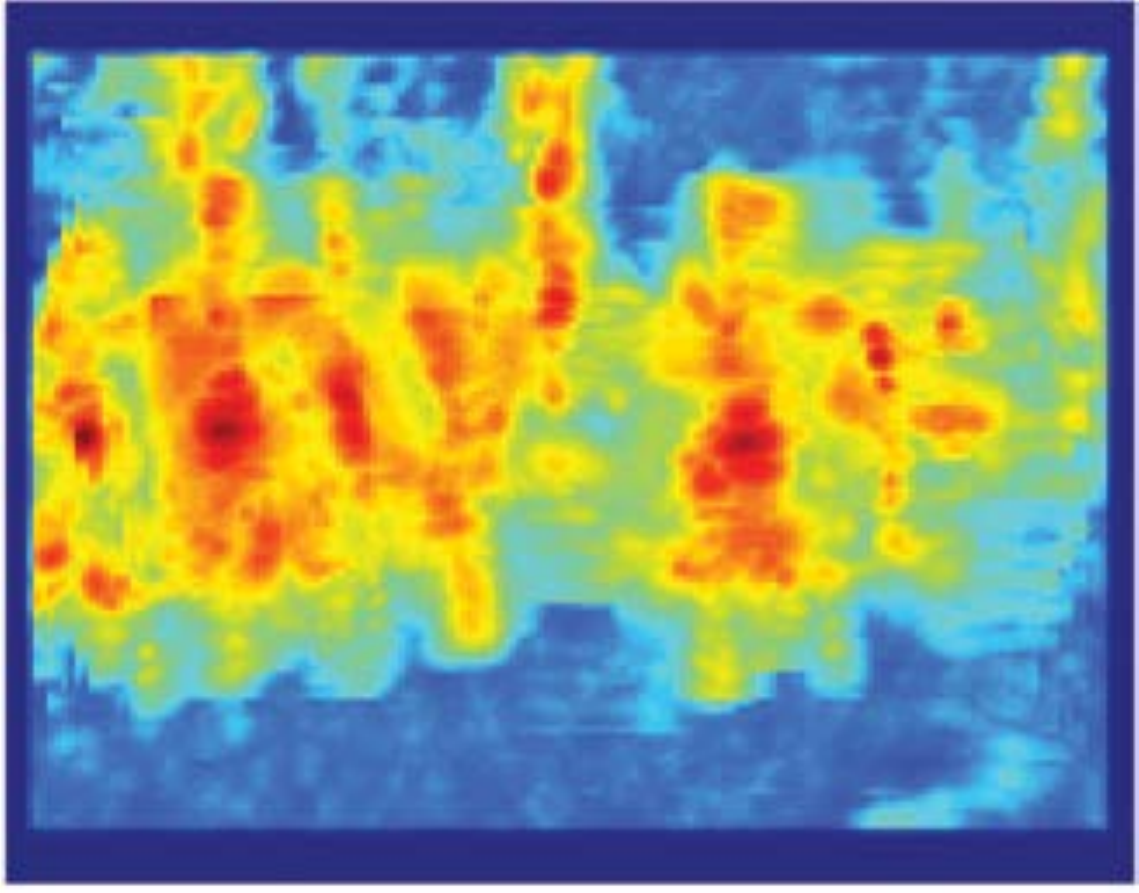}}
\subfigure[]{{}\includegraphics[width=0.24\linewidth]{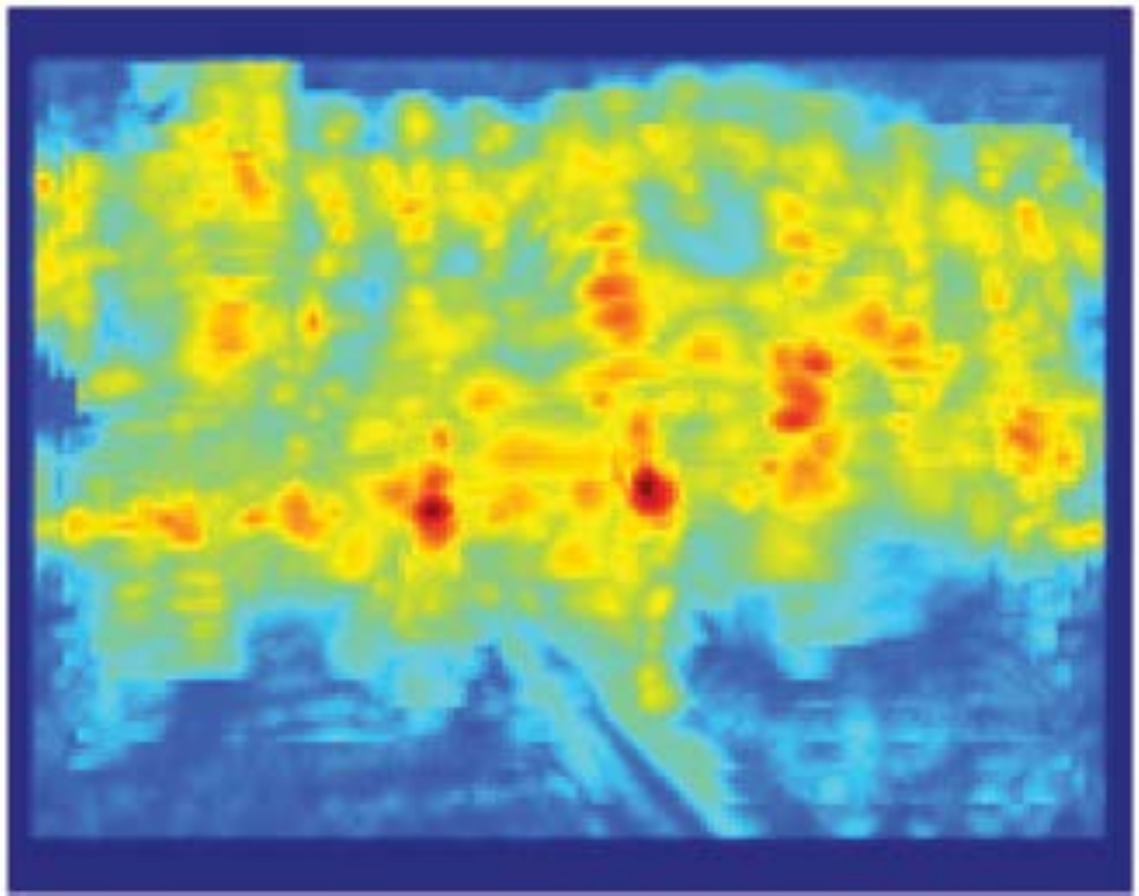}}
\end{center}
   \caption{LSI Results show multiscale detection of pedestrians across wide range of scales. The estimated likelihood of pedestrian's location measured across all the scales is shown under each frame. As before, the dark red to reddish black denotes high to very high confidence of detector.}
\label{F7B}
\end{figure*}
\begin{figure}
\begin{center}
\includegraphics[width=\linewidth]{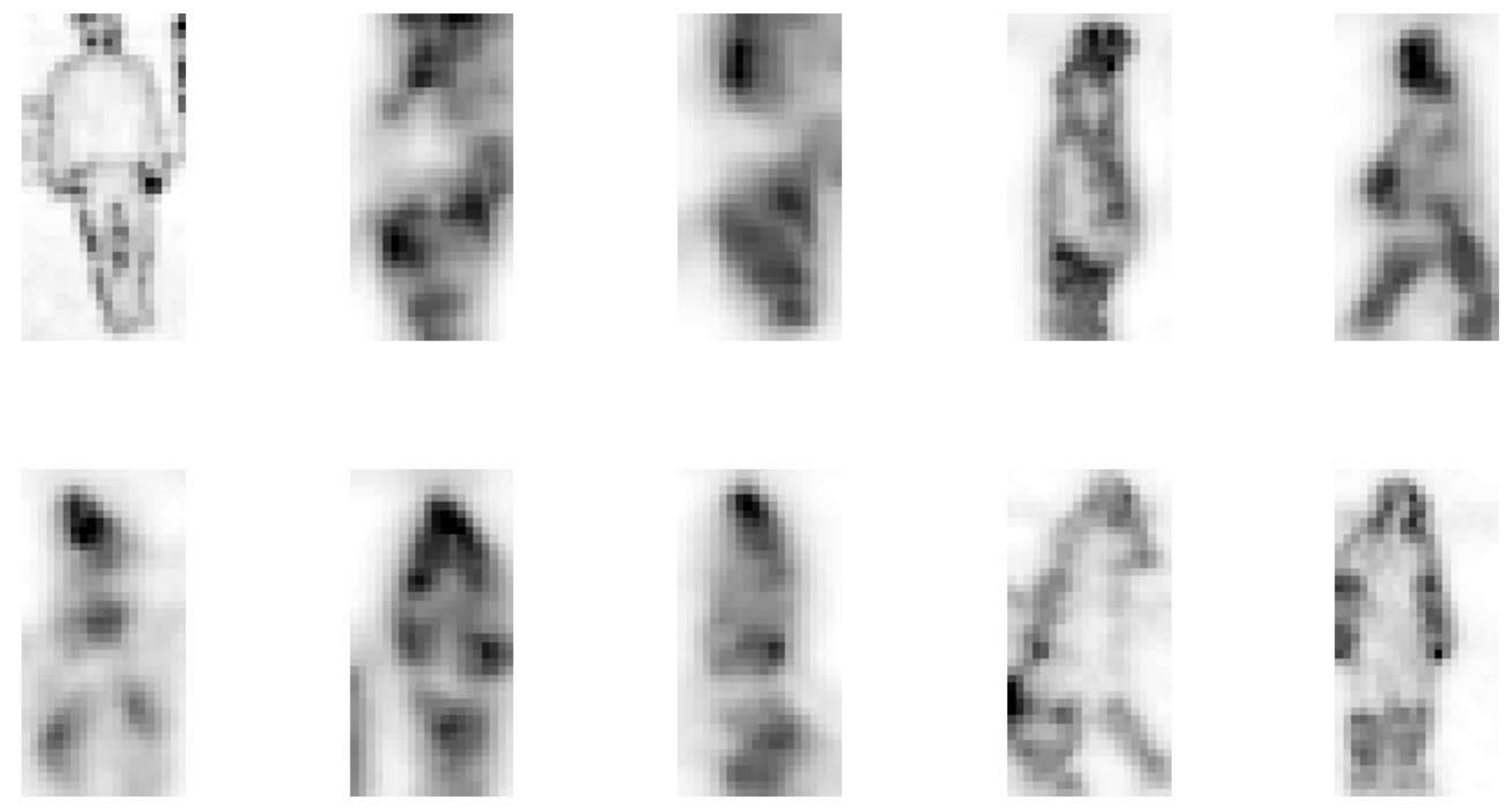}
\end{center}
\caption{The shape of pedestrian is prominent positive support tensors shown in the form of first LSK feature channel.  More importantly, the positive support tensors show how the linear kernel has succeeded to learn a set of widely different poses of pedestrians.}
\label{F7C1}
\end{figure}
\begin{figure}
\begin{center}
\includegraphics[width=\linewidth]{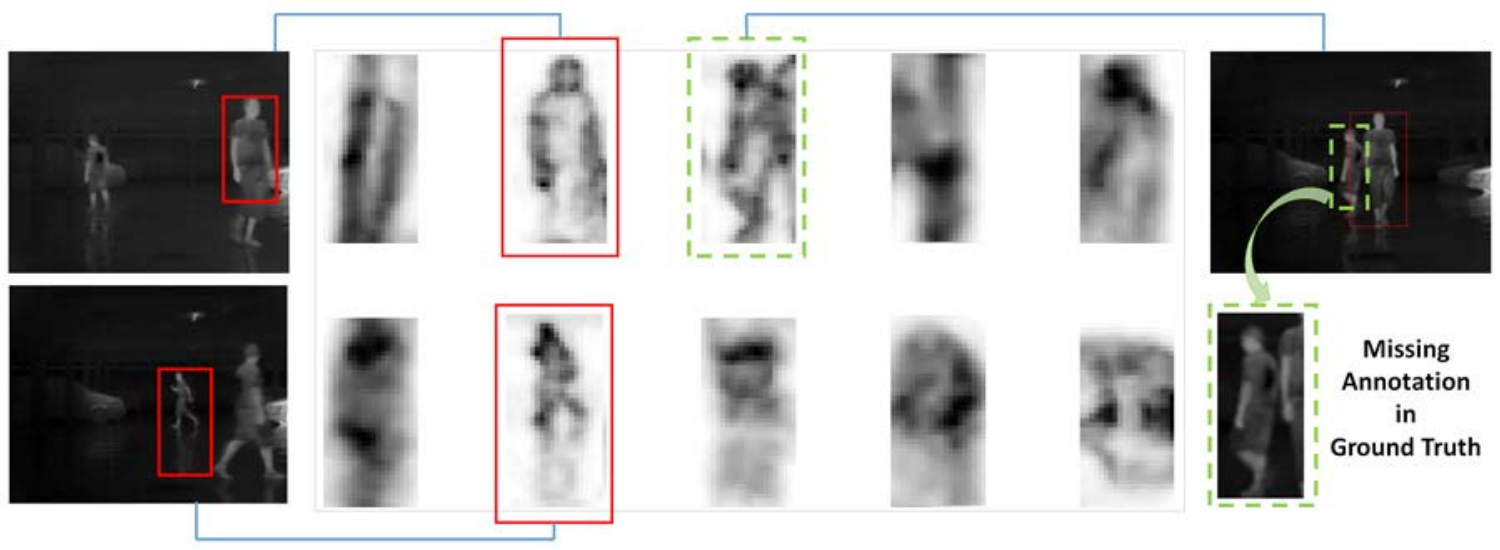}
\end{center}
\caption{Negative support tensors are shown to have come from hard mining step where undersized or oversized detection have resulted into false negatives (on left). On right, we show an instance where the correct detection is made but absence of such annotation in ground truth has forced this example into being a false negative.}
\label{F7C2}
\end{figure}

{\bf LSI Far Infrared Pedestrian Database}: This dataset comes in two flavors, classification setup as well as detection setup. We have focused on the detection set which is further divided in two subsets, training and test set. The training set has 3225 positive images and 1601 negative images. The test set includes 3279 positive and 4859 negative images. The images are 164 pixels wide and 129 pixels tall. Since the intensities of LSI images roughly range from 31000 to 35000 (16 bit images), we have scaled the intensities to 0-255 without noticeable loss in performance. We have followed the usual two-step process for the detector development: building an initial detector in the first step with a subset of positive and randomly sampled negative examples, and in the second step, we consider all positive examples besides including the hard negative examples from the initial detector's output. A tensor of size 40$\times $20 with 3 channels are learnt, and the number of scales in the feature pyramid (Fig. \ref{F3}) is set at ten, namely, 2.50, 1.58, 1.16, 0.90, 0.75, 0.64, 0.55, 0.49, 0.44, 0.40. We use an The $\alpha$ value of 0.4 in this experiment.

Fig. \ref{F7A} shows the LSK feature channels corresponding to a thermal image. The LSK features characteristically decomposes the gray scale image in contour, horizontal and vertical segments. In general, we have observed that the number of support tensors in the final model ranges from 15$\%$ to 20$\%$ of the full dataset. In Fig. \ref{F7C1} we show the positive support tensors by displaying the first channel of LSK tensor feature. One can notice the wide range of poses captured in the learning process of the support tensors. Fig. \ref{F7C2} illustrates negative support tensors which have resulted from hard mining step after being either under or over detected bounding box. The same figure also illustrates an example where missing annotation in ground truth pushes it into hard negative set. This shows that the proposed methodology is robust to noise and outliers present in the ground truth.
\begin{figure*}[t]
\begin{center}
\subfigure[]{{}\includegraphics[width=0.3\linewidth]{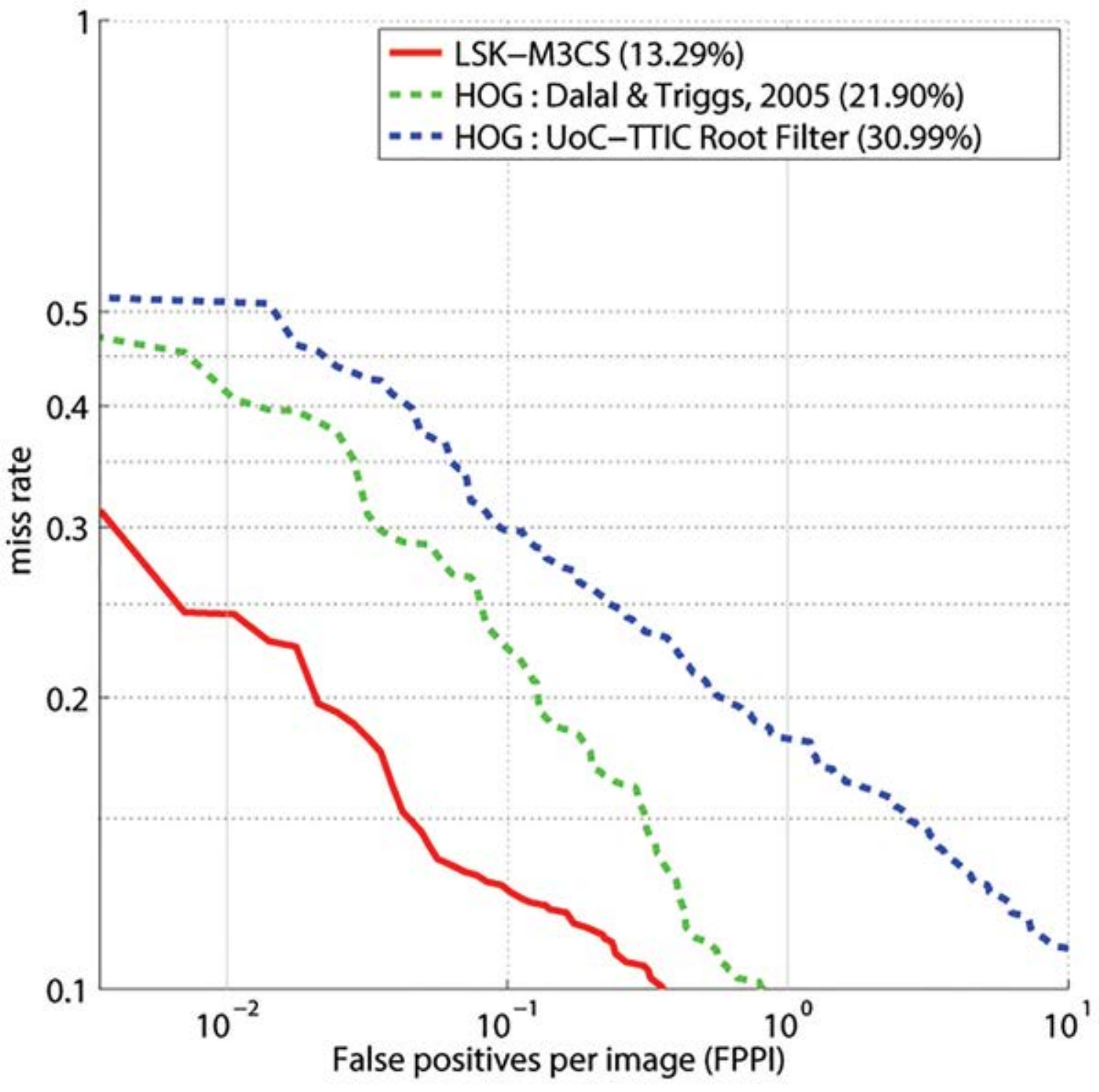}}
\subfigure[]{{}\includegraphics[width=0.3\linewidth]{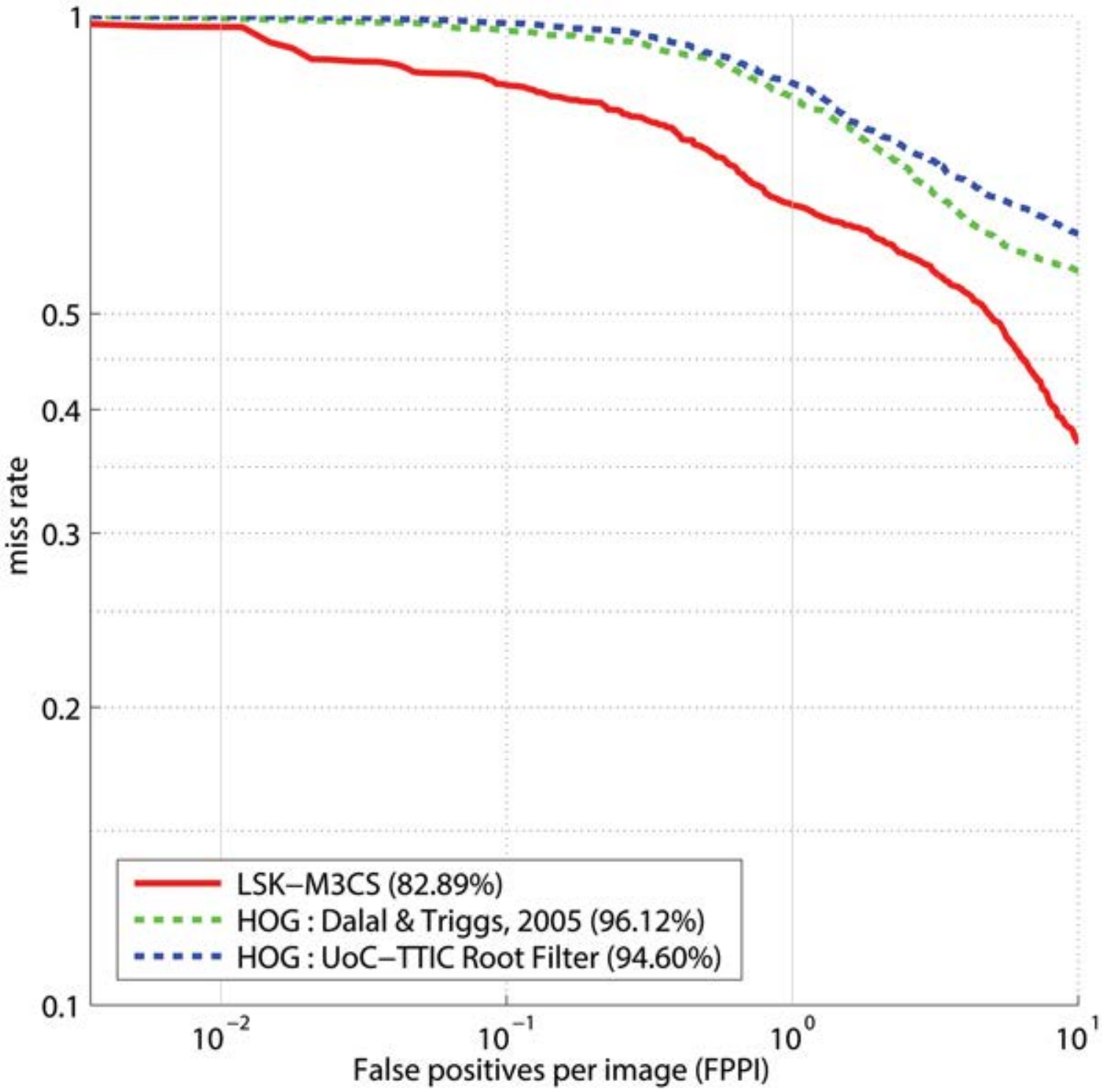}}
\subfigure[]{{}\includegraphics[width=0.3\linewidth]{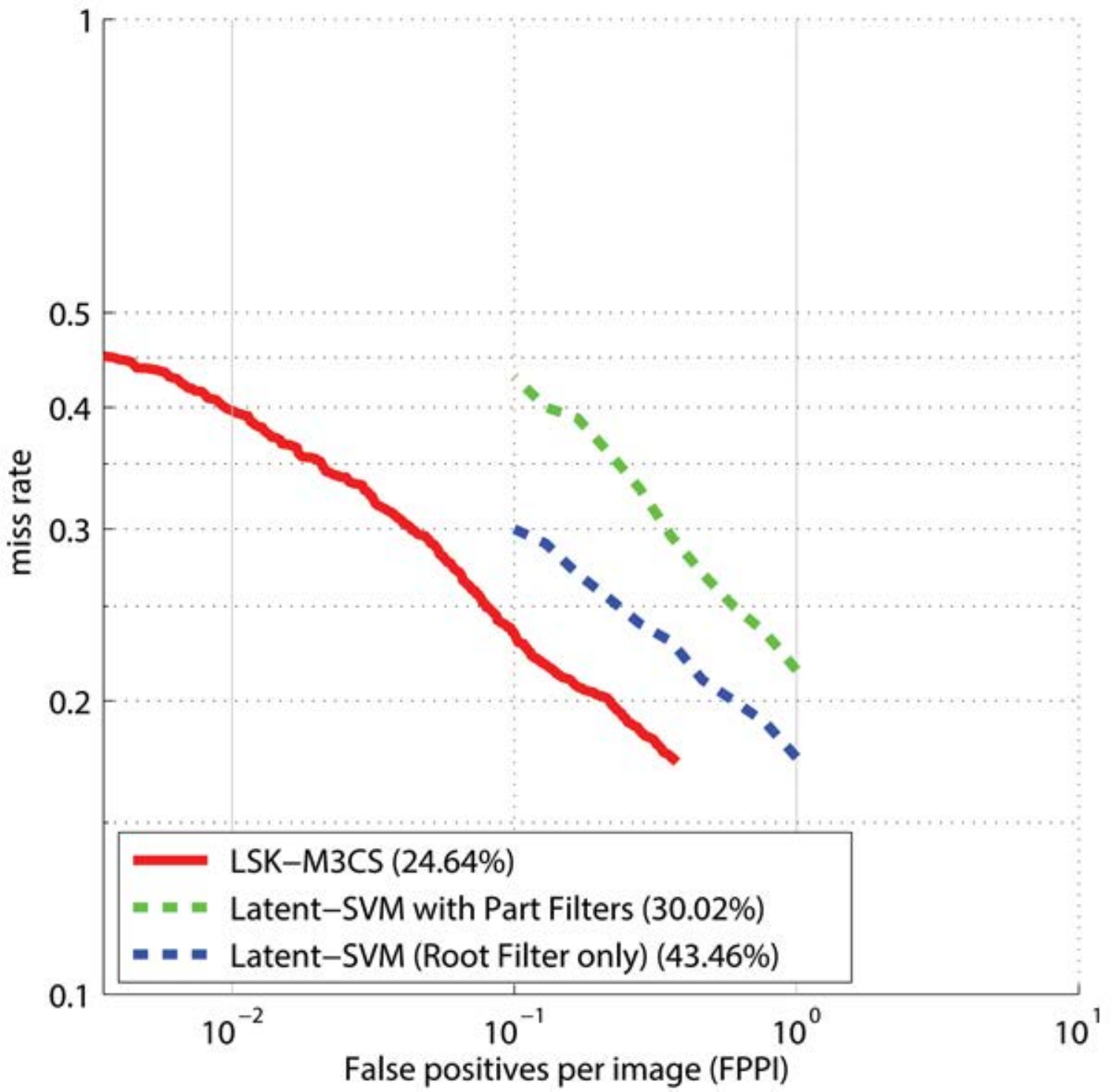}}
\end{center}
   \caption{Miss rate versus false positives per image (FPPI) for the three datasets: (a) OSU-T, (b) OSU-CT and (c) LSI thermal dataset. The miss rates at $10^{-1}$ FPPI are mentioned for the proposed LSK-M3CS and other baselines.}
\label{F8}
% use PrecisionW/osuT/manageEval.m to create figs0.
\end{figure*}
\subsection{Results \& Discussions}
\begin{table}
\caption{Runtime of Fast Object Detection with Scale Estimation in Comparison with Sliding Window Scheme}
\label{T0}
\begin{center}
\scalebox{0.85}{
\begin{tabular}{|c|c|c|c|c|c|}
\hline
{Datasets} & {Detector} & {Image} & \multicolumn{3}{|c|}{Detection time (in seconds)} \\ \cline{4-6}
{} & {height $\times$ width} & {height $\times$ width} & {LSK} & {HOG} & {HOG} \\
{} & {(pixels)} & {(pixels)} & {M3CS} & {(DT)} & {(UoC-TTIC)} \\ \hline
{OSU-T} & {36 $\times$ 28} & {240$\times$ 360} & {0.15} & {6.42} & {7.61} \\
{(single scale)} & {} & {} & {} & {} & {} \\ \hline
{OSU-CT} & {30 $\times$ 20} & {240$\times$ 320} & {0.28} & {24.01} & {26.41} \\
{(6 scales)} & {} & {(1.3 to 0.5)} & {} & {} & {} \\ \hline
{LSI} & {40 $\times$ 20} & {129$\times$ 164} & {0.31} & {-} & {-} \\
%{LSI} & {40 $\times$ 20} & {129$\times$ 164} & {0.31} & {38.24} & {40.18} \\
{(10 scales)} & {} & {(2.5 to 0.4)} & {} & {} & {} \\ \hline
\end{tabular}}
\end{center}
\end{table}
There are usually two approaches available when it comes to computing features, namely, feature engineering and feature learning. Boosting, sparse coding and recently convolutional neural network are learning methodologies one can apply for learning the features from raw image pixels. On the other hand, engineered features, especially HOG has dominated the object detection scenario in the first decade of this century leading to the success of several state of the art detectors, for example, deformable part model.

In our work, however, two of the datasets, namely OSU-T and OSU-CT have pedestrians so small that explicit modeling of parts is not a feasible idea to apply. We have implemented a HOG based linear SVM like \cite{dalal2005histograms} using the MATLAB library VLFeat \cite{vedaldi08vlfeat} as a baseline for comparison purpose. HOG implemented in VLFeat comes in two forms, one being the originally proposed feature in Dalal and Triggs, 2005 \cite{dalal2005histograms}, and the other is the dimension-reduced form used in \cite{felzenszwalb2010object} (denoted by UoC-TTIC in Fig. \ref{F8}).

Our proposed LSK with max-margin MCS kernel (LSK-M3SC) detector works superior to HOG based linear SVM both on OSU-T and OSU-CT achieving lowest miss rate (Fig. \ref{F8}(a) and (b) respectively). However, the extremely occluded nature of pedestrians in OSU-CT has made the detection task challenging for both HOG and LSK with MCS kernel. Fig. \ref{F8}(c) shows the performance of proposed detector on LSI dataset in comparison with HOG root filter and Latent-SVM with parts \cite{LSI,felzenszwalb2010object}. We refer the reader to \cite{LSI} where the authors have pointed out how introduction of parts in Latent-SVM introduces a derogatory performance on LSI as it often leads to some confusion of the part detector in absence of robust texture in the low resolution and noisy image environment. The proposed feature with our chosen MCS kernel has been able to achieve minimum miss rate as shown in Fig. \ref{F8}.

Owing to the efficiency of Fourier transform and integral image the detection process is pretty fast. We have conducted our experiment on a pretty standard Intel Xeon 64-bit desktop machine (CPU E3-1246 v3 @ 3.50GHz) with Ubuntu Linux 14.04 LTS. The performance results in terms of runtime are shown in Table \ref{T0} (DT\cite{dalal2005histograms} and UoC-TTIC \cite{felzenszwalb2010object} are two HOG implementations available in VLFeat). In the current implementation, most of the time is spent by the detector in feature computation. Therefore, we have accelerated the feature computation stage with an elementary C-mexfile implementation. The other modules (e.g., non-maximum suppression, scale estimation) of the proposed detection algorithm are implemented in MATLAB. It goes without saying that the HOG based template detection over six scales takes considerably longer time spanning few seconds to complete. The proposed fast detector also accelerates the training process by making the hard mining step quicker. The search of pedestrians over six scales in a typical 240$\times$ 360 (in OSU-T dataset) and 240 $\times$ 320 image (in OSU-CT dataset) happens in about a second. The detection in LSI is even faster because of its relatively smaller size 129 $\times$ 164.

\section{Conclusion}
In this paper we have extended and investigated the use of LSK tensors for pedestrian detection task in thermal infrared images. We have argued that when viewed in the lens of tensors, LSK offers many notable advantages like robustness, noise modeling, superior localization performance, and efficient detection. Continuing in this direction we have proposed a general framework for learning a tensor detector with Matrix Cosine Similarity, as a kernel function. The resulting maximum margin framework is able to distinguish pedestrians from background in challenging scenarios ranging from low signal images to detection at a far away distance. An exact acceleration of the classifier function is proposed by leveraging the tensor form of features as well as multi-channel signal processing techniques. The proposed methodology is compared with other state of the art detectors known to perform well with visible range sensors on the publicly available data sets of thermal infrared images.
\ifCLASSOPTIONcaptionsoff
  \newpage
\fi

\bibliographystyle{IEEEtran}
\bibliography{LSK_M3CS_bib_v1}
\end{document}